\documentclass[journal]{IEEEtran}
\usepackage{cite}

\ifCLASSINFOpdf
  \usepackage[pdftex]{graphicx}
  \usepackage{import}
  \graphicspath{{images/authors/}{images/figures/}}
  \DeclareGraphicsExtensions{.pdf,.jpg,.jpeg,.png,.pdf_tex}
\else
\fi
\usepackage{mathtools}
\usepackage{amsmath}
\usepackage{amssymb}
\usepackage{accents}
\usepackage{textcomp}
\usepackage{gensymb}
\usepackage{xfrac}
\DeclareMathOperator*{\argmin}{arg\,min}
\usepackage{amsfonts}

\usepackage{algorithmic}

\usepackage{array}
\usepackage{url}

\usepackage{color,soul}
\usepackage{xcolor}
\usepackage{threeparttable}
\usepackage{hyperref}

\begin{document}
%
\title{Road Curb Detection and Localization\\ with Monocular Forward-view Vehicle Camera}
%
%
%

\author{Stanislav~Panev, 
        Francisco~Vicente, 
        Fernando~De~la~Torre 
        and~V\'{e}ronique~Prinet
\thanks{S. Panev, F. Vicente, and F. De la Torre are with the
Carnegie Mellon University, Pittsburgh, PA 15213-3815 USA
(e-mail: spanev@cmu.edu; \mbox{franciscovicencar@gmail.com}; ftorre@cs.cmu.edu).}
\thanks{V. Prinet is currently with The Hebrew University of Jerusalem, \mbox{Israel}.
This work was initiated while she was at General Motors.
(e-mail: vprinet@gmail.com).}
\thanks{
	Manuscript received December 15, 2017;
	revised June 30, 2018 and September 15, 2018;
	accepted September 20, 2018.
	Date of publication November 12, 2018;
	date of current version August 27, 2019.}}

\IEEEpubid{
	\begin{minipage}[t]{\textwidth}\ \\[0pt]
		\centering{
		\copyright~2019 IEEE. Personal use of this material is permitted.
	Permission from IEEE must be obtained for all other uses, in any current or future media, including reprinting/republishing this material for advertising or promotional purposes, creating new collective works, for resale or redistribution to servers or lists, or reuse of any copyrighted component of this work in other works.	
	}
	\end{minipage}}


\maketitle

\begin{abstract}
We propose a robust method for estimating road curb 3D parameters (\textit{size, location, orientation}) using a calibrated monocular camera equipped with a fisheye lens. Automatic curb detection and localization is particularly important in the context of Advanced Driver Assistance System (ADAS), i.e.\ to prevent possible collision and damage of the vehicle's bumper during perpendicular and diagonal parking maneuvers.  
Combining 3D geometric reasoning with advanced vision-based detection methods, our approach is able to estimate the vehicle to curb distance in real time with mean accuracy of more than 90$\%$, as well as its orientation, height and depth.

Our approach consists of two distinct components -- curb detection in each individual video frame and temporal analysis. The first part comprises of sophisticated curb edges extraction and parametrized 3D curb template fitting. Using a few assumptions regarding the real world geometry, we can thus retrieve the curb's height and its relative position w.r.t. the moving vehicle on which the camera is mounted. Support Vector Machine (SVM) classifier fed with Histograms of Oriented Gradients (HOG) is  used for appearance-based filtering out outliers. In the second part, the detected curb regions are tracked in the temporal domain, so as to perform a second pass of false positives rejection.

We have validated our approach on a newly collected database of 11 videos under different conditions.
We have used point-wise LIDAR measurements and manual exhaustive labels as a ground truth.
\end{abstract}

\begin{IEEEkeywords}
curb detection, parking assistance, monocular camera, HOG, SVM, template fitting, tracking.
\end{IEEEkeywords}

%
\IEEEpeerreviewmaketitle

\section{Introduction}
\label{sec:introduction}
%
%
%
%

\IEEEPARstart{O}{ver} the last few years, the automotive industry has been focused on developing autonomous driving vehicles to reduce accidents and increase independence. As an intermediate step toward fully autonomous vehicles, the importance of active safety technologies, such as adaptive cruise control, blind spots warning, and automatic  park system, has  increased. Those features rely for most on sensor-based technologies, that try to understand the host vehicle's surrounding, i.e.\ to detect dynamic and static obstacles within a certain range. For example, moving objects, such as pedestrians and vehicles, can be detected to warn drivers to be cautious. Automatic detection of road signs can be used to control or adjust vehicles speed.

Curbs or sidewalks (in addition to road surface markings) are clues that are exploited in positioning systems.
Often they indicate the boundary of parking areas. Technologies that can accurately detect and estimate curb location and height are embedded in any assist/autonomous parking systems: they enable to predict the vehicle-to-curb distance, hence to avoid a potential collision between the curb and the vehicle, cause of damages on tires and bumpers. The constraints put on such systems are very high: near-zero false negative detection,  distance and height estimation with centimeter accuracy.

\IEEEpubidadjcol

The challenges to build such systems are at least two-fold. First, curbs are objects of small size. This compels to  use sensors of very high resolution to capture data where the object of interest covers a sufficiently large region. Second, curb shapes and appearance textures can vary drastically (depending on weather condition, pavement material, painting, etc.). This requires the development of advanced recognition techniques, capable to robustly classify an object as a curb or not.

\begin{figure}[t]
	\centering
	\def\svgwidth{\columnwidth}
	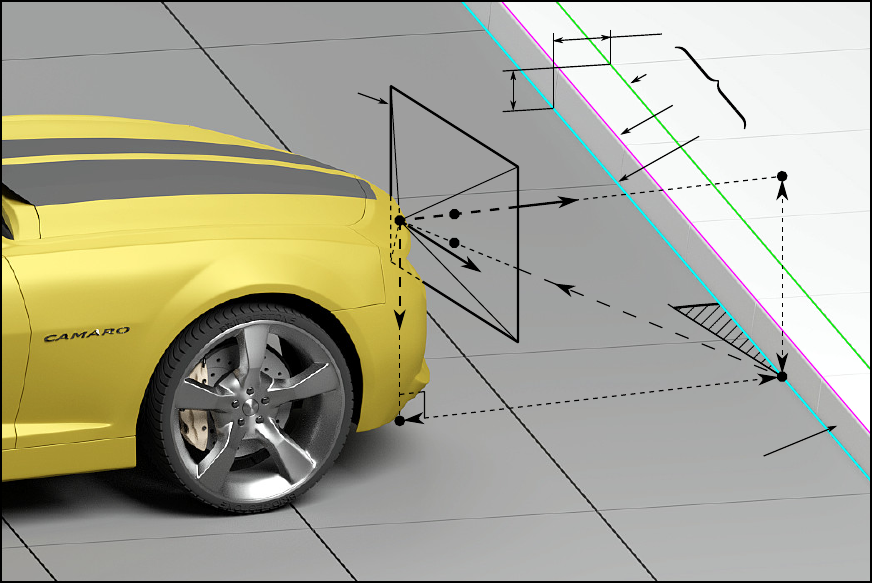
	\caption{Our system estimates the relative position, orientation and size of a curb w.r.t. a host vehicle,
		by the means of a monocular forward-viewing fisheye camera,
		advanced geometrical reasoning, temporal analysis and machine learning.}
	\label{fig:cameraPos}
\end{figure}

Most common techniques in the literature tackle mainly the first issue, i.e.\ 3D detection, using active sensors (LIDAR, laser range finders, etc.) or camera stereo-vision systems.  Those approaches assume that the curb's shape is in itself  a discriminative and robust feature. They are likely to fail in poor SNR conditions (e.g.\ bad weather) or when facing damaged curbs. To our knowledge, few work attempted to couple 3D geometry with  vision-based appearance models, so as to improve robustness and accuracy.

In this paper, we describe a system intended to assist the drivers and reduce the risk of running over obstacles, such as the curbs, in the everyday parking activities. Thereby, preventing unwanted damages and accidents.
A single monocular wide-angle camera is used as an input device,
which is one of the most economically efficient solutions nowadays.
The cost of such sensors sometimes can be several magnitudes lower than the price of the more complex devices, such as LIDARs and stereo cameras.
Our system works in real time, while maintaining high detection rate and curb parameters estimation accuracy. It relies on geometric reasoning, simple hand-crafted features (image edges and \textit{Histograms of Oriented Gradients} -- HOG), 
model-based machine learning techniques (\textit{Support Vector Machines} -- SVM) and temporal filtering. 
Our \textit{Inverse Perspective-compressing Mapping} (IPcM) technique approaches the curb edge detection in a sophisticated scale-invariant manner,
without the need of maintaining large multi-scale image space in order to reliably handle the broad variations of the curb size in the image.
All that allowed us to come up with an CPU-based implementation giving our system high level of flexibility.
For example, the current setup consists of a single front-mounted fisheye camera,
which is applicable to perpendicular and diagonal forward parking scenarios.
Just by attaching few more cameras to the system that cover the lateral and rear perimeters around the vehicle,
will extend the system applicability to perpendicular reverse and in-line parking.
The aforementioned advantages also would let our system to operate on a single computing platform in conjunction with algorithms designed to solve more challenging tasks based on deep models, such as motion planning and control.
Thus, each subsystem occupies separate processing unit - CPU and GPU.

\section{Related Work}

The research of the curb detection algorithms can be figuratively organized according to the types of the sensors employed.
Undoubtedly, the LIDARs are among the most popular active sensors. They provide 3D point cloud data based on laser scanning and range measurements.
In \cite{zhao_curb_2012}, for example, the authors voxelize the LIDAR data for computational efficiency and detect those containing ground points, based on the elevation information and plane fitting. The candidate curb points are selected using three spatial cues. Employing short-term memory technique along with a parabolic curb model and RANSAC they remove the false positives. For temporal curb tracking a particle filter is used. In \cite{yao_road_2012} the curb is modeled as parabola and Integral Laser Points features are used for speed up. Instead of temporal filters and spline fitting methods, in \cite{hata_robust_2014} a robust regression method to deal with occluding scenes, called Least Trimmed Squares (LTS), is used in combination with Monte Carlo localization.
In \cite{chen_velodyne-based_2015} instead of extracting features, the LIDAR scan lines are processed directly. Initial curb point candidates are determined by Hough transform and then iterative Gaussian Process Regression is used to represent the curb models.
In \cite{zhang_real-time_2015} the parabola model is employed as well, but the tracking technique is based on Kalman filter in combination with GPS/IMU data.

Another active sensor which provides 3D point cloud data is the Time-of-Flight (ToF) camera. It extracts the depth information from a 3D scene based on the phase shifts of light signals caused by the different times they travel in space to bounce off the objects and return back to the camera.
In \cite{scheunert_free_2007} the authors take advantage of the ToF camera's high frame rate to improve the results by space-time data accumulation using grid based approach. For estimating ego-motion parameters they employ Kalman filter. In \cite{gallo_robust_2008} CC-RANSAC method is used for improved plane fitting into the raw point cloud data.

The laser range finders (LRF) are active sensors from the LIDARs family, but instead of providing 3D point cloud data, they usually scan just a single line and estimate the distances of each measurement point along it. The curb detection algorithm in \cite{byun_autonomous_2011} is accomplished in two steps. Firstly, the authors detect the potential curb positions in the LRF data, then they refine the results by employing Particle filter.
In \cite{liu_curb_2013}, LRF data captured sequentially is used to build local Digital Elevation Maps (DEM) and Gaussian process regression is at the final curb detection stage.
A set of 3 LRF sensors is used in \cite{pollard_step_2013}. Peak detection is accomplished on the results of derivative-based distance method described there and then they merge the data from the individual sensors.

A popular passive sensor is the stereo camera. Similar to the LIDARs, it provides 3D point cloud data which has higher resolution, but usually contains more noise.
DEMs are often used for efficient representing the 3D data in the area of curb detection. In \cite{oniga_curb_2008}, edge detection is applied to the DEM data to highlight the the height variations. The noise from the stereo data is reduced significantly by creating multi-frame persistent map. Hough accumulator for lines is built with the persistent edge points. Each curb segment is refined using the RANSAC approach to fit optimally the 3D data of the curb.
In \cite{siegemund_temporal_2011}, a 3D environment model is utilized. It consists of various primal entities, such as road, sidewalk, etc. The 3D data points are assigned to the different part of the model using temporally integrated Conditional Random Fields (CRF).
In \cite{oniga_curb_2011}, temporal integration of the DEM data is also used, but in combination with least squares cubic spline fitting. 
The algorithm described in \cite{kellner_multi-cue_2015} presents an interesting idea of combining the 3D point cloud data with the intensity information from the stereo camera. First, they extract model-based curb features from the 3D point cloud and validate them by using the intensity image data. The curbs are presented as 3D polynomial chains.
The approach described in \cite{fernandez_curvature-based_2015} is based on the curvature of the 3D point could data. It is estimated by applying the nearest neighbor paradigm. The method's performance is evaluated by applying it to both -- stereo camera and LIDAR data.

All the sensor types used in the algorithms above directly provide some kind of 3D information, either point cloud or line-wise. However, monocular cameras data lacks completely from depth information. Thus, extracting curbs using them is a challenging task, usually founded on preliminary constraints and assumptions.
In \cite{haltakov_scene_2012} the image is divided in regular grid of cells. Then 3D reconstruction is applied by pixel-wise image labeling based on CRFs.
Besides the camera, the vehicles CAN-bus data is employed as well in \cite{seibert_camera_2013}. Then two complementing methods are applied:
localizing borders using texture based area classification with local binary patterns (LBP) features and Harris features tracking using Lucas-Kanade tracker to extract 3D information.
The curb detection system described in \cite{prinet_3d_2016} is closely related to our approach. It also involves use of a fisheye camera and incorporates Histogram of Oriented Gradients (HOG) features. Unlike our approach, they preserve the original camera image. Hence, their curb model is polynomial. The temporal filtering is based on Kalman filter. 

\section{System Description}
\label{sec:systemDescription}

\subsection{Algorithm summary}


The objective of our system is to acknowledge the presence of a curb close to the vehicle,
along its forward motion path, and to help identifying if it is an immediate thread to the vehicle's integrity by estimating
its position, orientation and size relative to the vehicle.
These parameters constitute the system state vector $\mathbf{x}$, described later in Section~\ref{sec:assumptions}.
The system tracks just one curb at a time, as only the closest to the vehicle one is significant.

\begin{figure}[t]
	\centering
	\def\svgwidth{0.95\columnwidth}
	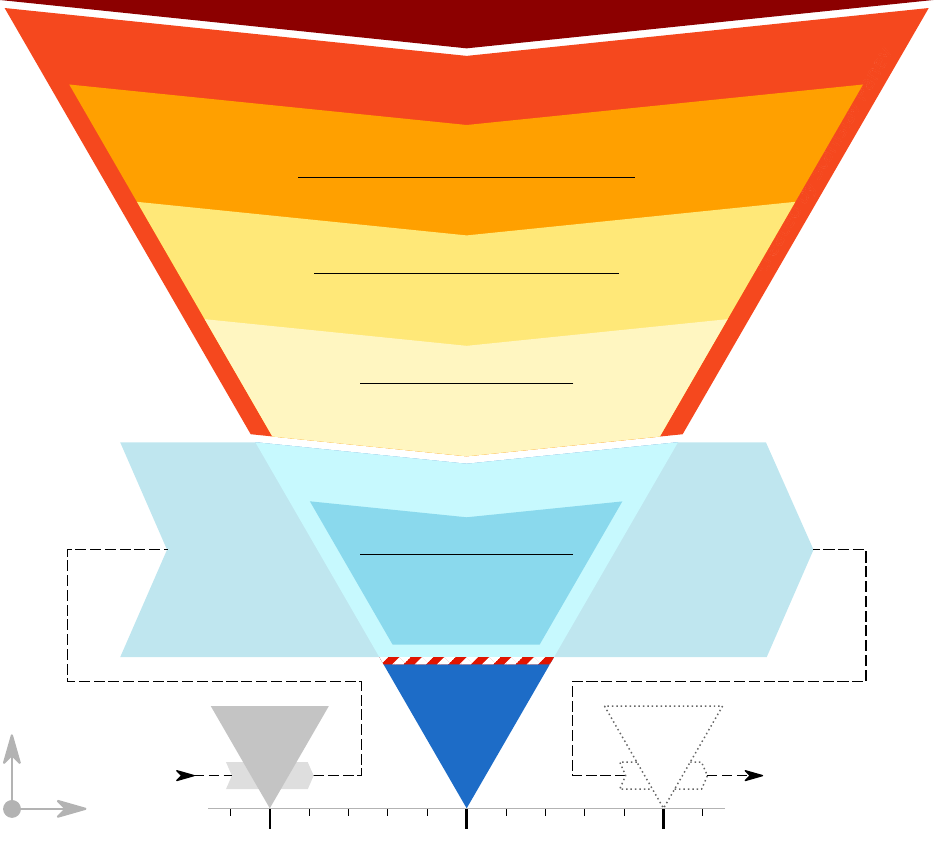
	\caption{System paradigm: the intersection between the spatial and temporal information cues is used ot minimize the uncertainty of curb parameters estimation.}
	\label{fig:algorithmParadigm}
\end{figure}

Fig.~\ref{fig:algorithmParadigm} illustrates graphically the paradigm our system is founded on.
Its shape likens inverse pyramid, situated in the \textit{Time}/\textit{Entropy} plain.
Pyramid tip points to the moment of time $t$ which corresponds to the last captured camera frame (image).
The width of its layers (and their coloring) depicts overall entropy rate in terms of the state vector $\mathbf{x}$ estimation.
The paradigm is inspired by the idea of the attentional cascade presented in \cite{ViolaJones}, but instead of boosted classifiers we use various filtering techniques.
The direction of processing flow is from the top to the bottom and each stage is purposed to reduce system's state entropy until
the uncertainty is low enough that a reasonable inference for the values of curb parameters can be made.

The cascade consists of two major layers which handle the information from space and time perspectives.
Immediately after a new image is delivered by the camera at time $t$, it is fed to the \textbf{``Spatial domain"} pyramid layer, which consists of three sub-layers.
The first one searches for individual \textit{curb's primitive structural elements} in the image. As such, we engage curb's edges,
since they are 3D lines. Projecting them onto the camera's image plane won't take away their straightness, because of the linearity of the perspective transform.
Therefore, curb's edges can be detected just by performing line detection in the image. This part of our algorithm is described in Section~\ref{sec:curbEdgeDetection}. 

Next, we raise the level of generalization up by utilizing curb's geometry itself, i.e.\ the configuration of its primitive structural elements (points, edges, faces) that constitute its 3D structure. Here, our algorithm relies on the prior knowledge of curb's geometry and the \textit{3D to 2D correspondence} in order to estimate which compositions of image lines could probably represent a projection of a 3D curb-like body. All the successful guesses mold the initial hypothesis set of \textit{curb candidates}.
Its outlying members are meant to be rejected by the next layers of our paradigm pyramid. Detailed description is presented in Section~\ref{sec:curbParamsEst}.

The last operation in the spatial domain shifts the focus from curb's geometry to its \textit{appearance} in the image.
Here we perform object detection to validate every curb candidate by the means of sophisticated Machine Learning techniques. More information can be found in Section~\ref{sec:curbCandidatesFiltering}.

The purpose of our system is to estimate curb parameters while vehicle is in motion. Luckily, it is a considerably massive object with predictable kinematics. Hence, the evolution of curb's parameters (system state) in the \textbf{``Time domain"} follow smooth trajectories, with no abrupt discontinuities. In other words, our system deals with environment which obeys \textit{temporal continuity}.
Thus, from all the curb candidates we can select only those which comply with it and also make reasonable predictions for system's future states.
Our Curb tracking technique is described in Section~\ref{sec:curbTracking}.

The most bottom layer of the pyramid represents the residual uncertainty which our system, as a non-ideal one, cannot resolve and bring the entropy to the theoretical value of 0, i.e.\ 100\% confidence about curb parameters estimates.
Our goal is to minimize it and in Section~\ref{sec:experimentalResults} we present results which demonstrate the promising performance of our system.


\subsection{Geometrical considerations and definitions}
\label{sec:assumptions}

Here we describe the fundamental assumptions and constraints
our algorithm is based on.

\subsubsection{Road} We consider the road as a perfectly flat structure.
Although, the real roadways are not ideally planar due to technological slopes or deformations caused by exploitation,
their surface curvature is smooth enough to allow us make such an assumption considering the size of our system's working area (\textit{Curb Detection Domain}, defined below in \ref{sec:CDZandRoI}) and the amount deviation from the perfect plane within it. The road plane is denoted as $\mathbf{\Pi}_{\mathrm{G}} \in \mathbb{R}^3$
and its projection in camera image as $\mathbf{\pi}_{\mathrm{G}} \in \mathbb{R}^2$ (Fig.~\ref{fig:cameraPos}).

\subsubsection{Camera} The camera is mounted in the middle of vehicle's front and points upon vehicles forward movement direction.
Camera's projection center (focal point) $O_\mathrm{C}$ is elevated above the road at a fixed distance $H_\mathrm{C}$
and this is where camera's coordinate frame $O_\mathrm{C} x_\mathrm{C} y_\mathrm{C} z_\mathrm{C} $ originates (Fig.~\ref{fig:cameraPos}). 
During parking vehicle's speed is relatively low, therefore we can neglect the actuation caused by vehicle's suspension system.
Furthermore, for the sake of simplicity and computational efficiency we define that camera's plane $x_\mathrm{C} z_\mathrm{C}$ is parallel to the road plane $\mathbf{\Pi}_{\mathrm{G}}$
(Fig.~\ref{fig:cameraPos}).

\subsubsection{Curb}
\label{sec:Curb}

\begin{figure}
\centering
\def\svgwidth{\columnwidth}
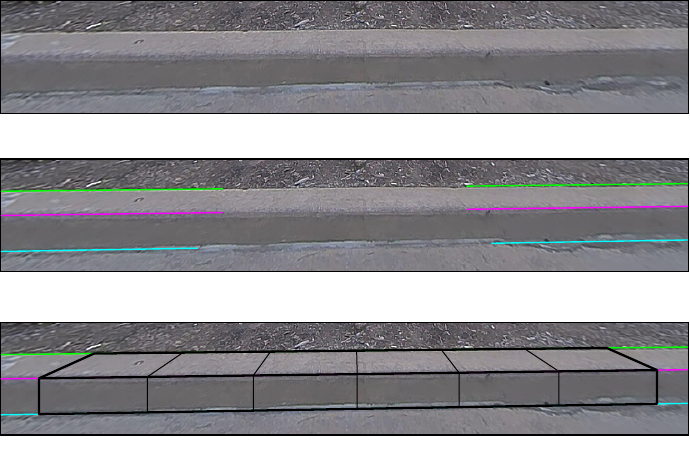
\caption{Curb elements, shape and definitions.}
\label{fig:curbDefinitions}
\end{figure}

We define the curbs as rectangular prismoidal rigid structures, which determine road plane borderlines. They are usually significantly elevated above the road surface and often specify the boundaries between the road and sidewalk, for example.
We also assume that the curbs have negligibly small fillets (roundings) of the corners, which results in clear and abrupt brightness transitions (edges)
in the camera images.

Our algorithm uses curb's edges as primal features.
They are straight, easily detectable and can help us to drastically reduce curb detection time.
We would like also to note that if a curb appears in the image, three of its horizontal edges and two faces defined by them will always be presented in it (Fig.~\ref{fig:curbDefinitions}a and b). Let $\mathbf{E}_1$, $\mathbf{E}_2$ and $\mathbf{E}_3 \in \mathbb{R}^3$ be the three lines depicting curb's lower front (base), upper front and rear edges, respectively (Fig.~\ref{fig:cameraPos}).
Accordingly, $\mathbf{e}_1$, $\mathbf{e}_2$ and $\mathbf{e}_3$ are their $\mathbb{R}^2$ projections in the image (Fig.~\ref{fig:curbDefinitions}b). 

Curb detection and localization aim to the estimation of four basic curb parameters: $D_\mathrm{U}$ -- the distance between the curb and camera, 
$\Theta_\mathrm{U}$ -- the rotation angle of the curb about vertical axis, $H_\mathrm{U}$ and $E_\mathrm{U}$ -- curb's height and depth, respectively.
To precise the definition of $D_\mathrm{U}$ we introduce the notion of curb's reference point $\mathbf{P}_\mathrm{U}$ (Fig.~\ref{fig:cameraPos}), which is the intersection between the plane $y_\mathrm{C} z_\mathrm{C}$ and $\mathbf{E}_1$ -- \textit{curb's base edge}. Then $D_\mathrm{U}$ can be described as the distance between $\mathbf{P}_\mathrm{U}$ and the orthogonal projection of the origin $O_\mathrm{C}$ on the road plane $O_\mathrm{C}^{\{\mathrm{G}\}}$. As a consequence of our system's simplified geometrical configuration (see above), $D_\mathrm{U}$ is equal to the $z$-coordinate of $\mathbf{P}_\mathrm{U}$ in camera's frame $O_\mathrm{C} x_\mathrm{C} y_\mathrm{C} z_\mathrm{C}$.
$\Theta_\mathrm{U}$ can be defined as the angle between camera's axis $x_\mathrm{C}$ and the edges $\mathbf{E}_{1,2,3}$,
$H_\mathrm{U}$ is the distance between $\mathbf{E}_1$ and $\mathbf{E}_2$, respectively, $E_\mathrm{U}$ is the distance between
$\mathbf{E}_2$ and $\mathbf{E}_3$ (Fig.~\ref{fig:cameraPos}). 
Semantically, we split curb's parameters in two groups -- essential and secondary.
$D_\mathrm{U}$, $\Theta_\mathrm{U}$ and $H_\mathrm{U}$ are considered as essential, 
because they provide enough information to determine the safe-clearance between the curb and vehicle.
$E_\mathrm{U}$ is considered secondary and it is used for an additional information cue to improve system's reliability during operation.

We define system's state vector as follows 
\begin{equation}
\label{eq:curbParams}
  \mathbf{x} = \begin{bmatrix} D_\mathrm{U} , \Theta_\mathrm{U} , H_\mathrm{U} , E_\mathrm{U} \end{bmatrix}^\top,
\end{equation}
and it holds all the curb parameters.
Their estimation is accomplished by the means of a 3D parametric template (shown in Fig.~\ref{fig:curbDefinitions}c).  
Similar to the curb, it consists of two orthogonal rectangular faces and three edges $\widehat{\mathbf{E}}_1$, $\widehat{\mathbf{E}}_2$ and $\widehat{\mathbf{E}}_3$
which correspond to the curb's ones. Consequently, their projections in the image plane are $\widehat{\mathbf{e}}_1$, $\widehat{\mathbf{e}}_2$ and $\widehat{\mathbf{e}}_3$, respectively.
The template has the same set of parameters as the curb and they are organized in the template's state vector
\begin{equation}
  \label{eq:curbTemplateParameters}
  \widehat{\mathbf{x}} = \begin{bmatrix} \widehat{D}_\mathrm{U} , \widehat{\Theta}_\mathrm{U} , \widehat{H}_\mathrm{U} , \widehat{E}_\mathrm{U} \end{bmatrix}^\top.
\end{equation}
Detailed description of the fitting procedure is presented in Section~\ref{sec:curbParamsEst}. 

A curb detection in the image frame at time $\mathit{t}$ is considered as successful, if at least its essential parameters are estimated correctly.
Therefore, we need to determine at least the position and orientation of $\mathbf{E}_1$ and $\mathbf{E}_2$ w.r.t. the camera from their projections $\mathbf{e}_1$ and $\mathbf{e}_2$, respectively.


\subsubsection{Curb detection domain and image's curb searching region}
\label{sec:CDZandRoI}

\begin{figure}[t]
\centering
\def\svgwidth{\columnwidth}
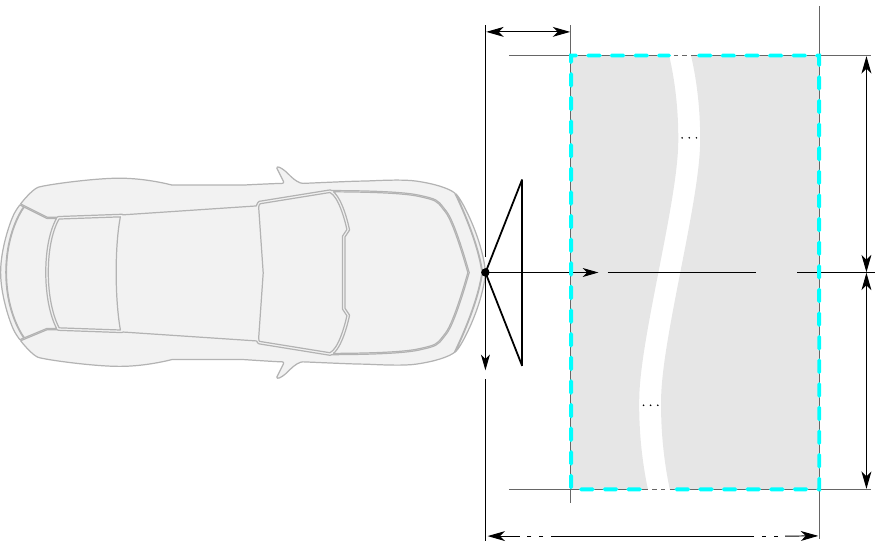
\caption{
  System's \textit{Curb Detection Domain} (CDD) \textit{(cyan)} -- shape and dimensions. 
}
\label{fig:curbDetectionDomain}
\end{figure}

From a practical point of view, we define the rectangular area of the road plane directly in front of the vehicle as \textit{Curb Detection Domain} (CDD)
(Fig.~\ref{fig:curbDetectionDomain}). In essence, CDD determines system's domain of definition (or operation) w.r.t. $D_\mathrm{U} \in [D_\text{min}, D_\text{max}]$.
$D_\text{max}$ is chosen to be 500 cm, whereas $D_\text{min}$ is calculated from camera's vertical \textit{Field of View} (FoV) bottom boundary.
In our setup its value is ${\approx}28.7$ cm.
The CDD's side limit $W_\text{max}$ is derived from FoV as well and is rounded to 130 cm.
The total area covered by the our system's CCD is ${\approx}12.25$ $\text{m}^2$.
The four sides of CDD are defined by the lines $\mathbf{B}_\mathrm{N}$, $\mathbf{B}_\mathrm{F}$, $\mathbf{B}_\mathrm{L}$ and $\mathbf{B}_\mathrm{R}$. Their projections in the image are respectively $\mathbf{b}_\mathrm{N}$, $\mathbf{b}_\mathrm{F}$, $\mathbf{b}_\mathrm{L}$ and $\mathbf{b}_\mathrm{R}$.

\begin{figure}
\centering
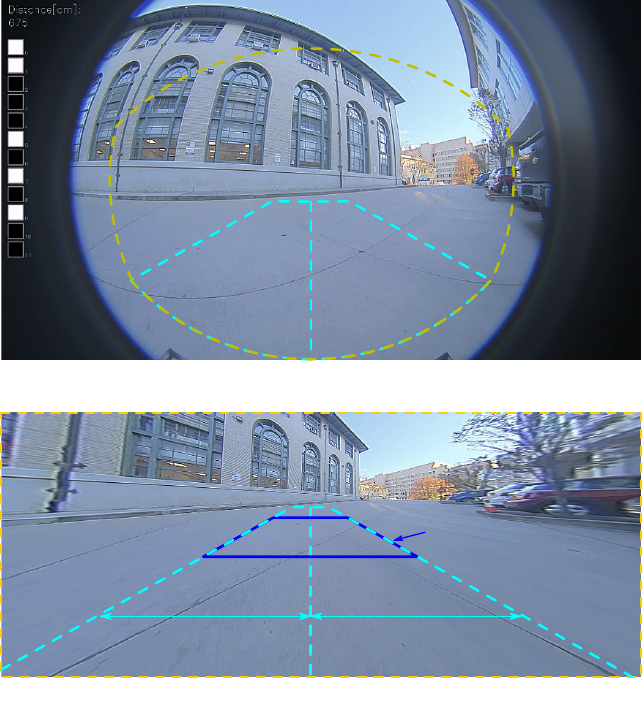
\caption{The original (a) and rectified (b) versions of an image from the forward-view vehicle fisheye camera.
  \textit{cyan dashed line} -- Curb Detection Domain (CDD), \textit{orange dashed line} -- boundaries of the rectified camera image,
  \textit{blue solid line} -- exemplar position of Curb Searching Region (CSR).} 
\label{fig:cameraView}
\end{figure}

In order to reduce the amount of data being processed for each frame and eliminate the influence of outliers,
we introduce the notion of \textit{Curb Searching Region} (CSR) in the image (Fig.~\ref{fig:cameraView}b).
It defines the image area used for extracting curb features.
Its size and position are variable and determined by the expected curb location at the time of current camera frame.
Essentially, it represents the projection in the image of a CDD subregion which has the same width,
but significantly shorter length. Its initial position is set at the far end of CDD and
in \textit{Curb tracking} mode (see Section~\ref{sec:curbTracking}) the system updates its position accordingly.

\subsection{System calibration}

\begin{figure}
	\centering
	\def\svgwidth{\columnwidth}
	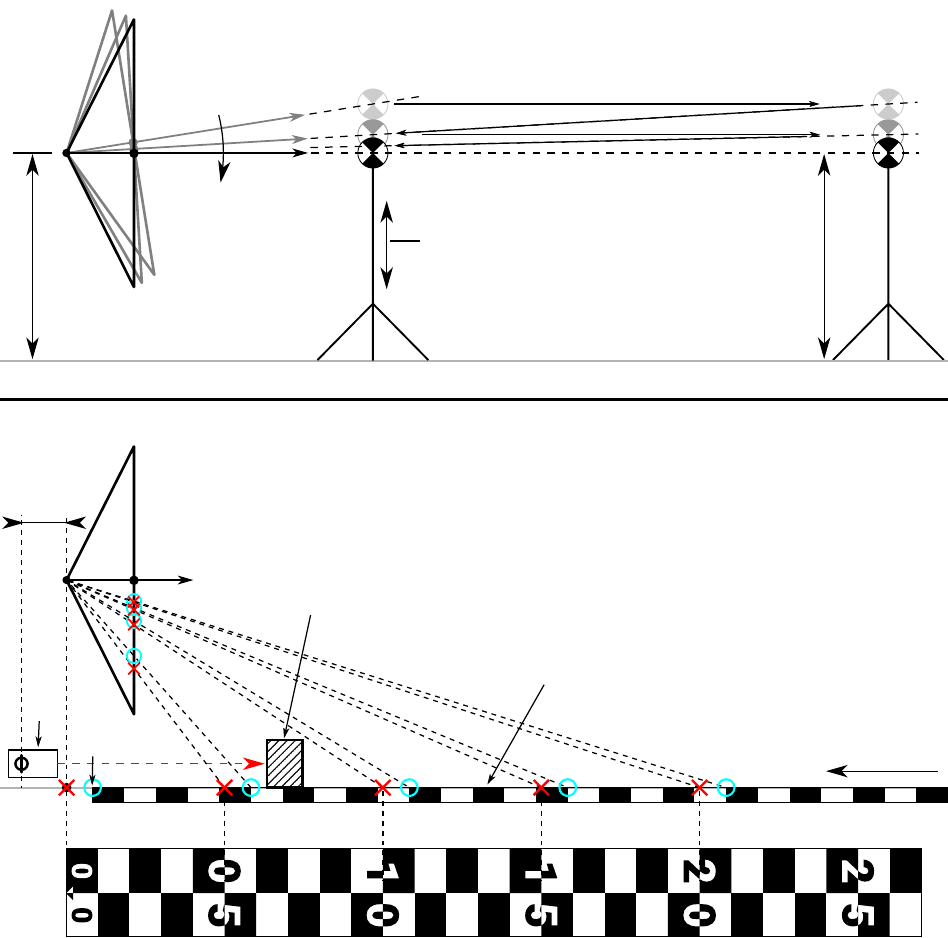
	\caption{System geometry calibration: (\textit{top}) Assure that camera optical axis is parallel to the road plane by the means of point calibration target with adjustable height. This procedure give us also an accurate estimate of camera elevation $H_C$.
	(\textit{bottom}) Estimate the distance offset $\Delta$ between the camera and LIDAR measurements.}
	\label{fig:geometryCalibration}
\end{figure}

All images from the camera are rectified before any further processing to eliminate
the radial distortions introduced by the fisheye lens (Fig.~\ref{fig:cameraView}).
Otherwise, curb detection would be much more complex,
involving second or higher order curves detection and fitting.
We employ the fisheye camera model described in \cite{fisheyeCalibration} to estimate camera's intrinsic parameters in an offline 
calibration procedure using a planar target.
In the rest of this paper, by the notion ``image" we refer to the rectified version of the original image,
unless anything else is explicitly stated.

The next step is, assuring that the camera extrinsic parameters follow the geometric definitions above (Section~\ref{sec:assumptions}).
We don't estimate those parameters through a calibration procedure.
Instead, we set some of them manually. For instance, camera tilt angle should be set to zero.
To achieve that, we employ a simple four-steps calibration procedure,
illustrated in Fig.~\ref{fig:geometryCalibration}-\textit{top}.
A point target (marker) mounted on a stand (tripod), whose height is adjustable, is used.
Repeating the these steps 3-4 time ensures that camera orientation will easily converge to the desired state.
Not only the correct orientation of the camera is set during this process,
but we also get an accurate estimate of the distance $H_C \rightarrow \widehat{H}_C$,
which is the distance from target center to the ground and can be measured with a ruler.

Fig.~\ref{fig:geometryCalibration}-\textit{bottom} illustrates our technique to estimate the horizontal offset between camera and LIDAR origins -- $\Delta$, which is needed when evaluating system distance measurement capabilities.
The peculiarity here is that the position of camera coordinate frame origin $O_{\mathrm{C}}$ is hard to be measured.
Therefore, we came up with a technique to estimate its position implicitly. More precisely, we do not measure its actual position, but the position of its projection on the road plane $O_{\mathrm{C}}^{\mathrm{\{G\}}}$. Thanks to our geometric setup this is sufficient to estimate $\Delta$.
We use a 3D model of a planar large-scale ruler with a regular grid (10 cm per division). First, we create a virtual 3D model of it, which is coplanar with the road plane $\mathbf{\Pi}_\mathrm{G}$, its origin coincides with $O_{\mathrm{C}}^{\mathrm{\{G\}}}$ and 
its measurement axis is parallel to $z_{\mathrm{C}}$.
Then we project this virtual model of the ruler to camera plane and overlay its projection
in the image.
Next, we take a printed version of the same ruler model with the same scale, lay it in front of the camera and fit it to the overlaid projection of its virtual analog.
When finished, we know that the origin of the printed ruler corresponds to the origin of the virtual one and, consequently, to $O_{\mathrm{C}}^{\mathrm{\{G\}}}$.
Afterwards, we use a test obstacle to initiate measurements from the LIDAR and the camera, through the ruler (see Section~\ref{sec:measureDistances}). By averaging the differences of those pairs of measurements we can calculate $\Delta$. 

\subsection{Measuring distances with a monocular camera}
\label{sec:measureDistances}

\begin{figure}
\centering
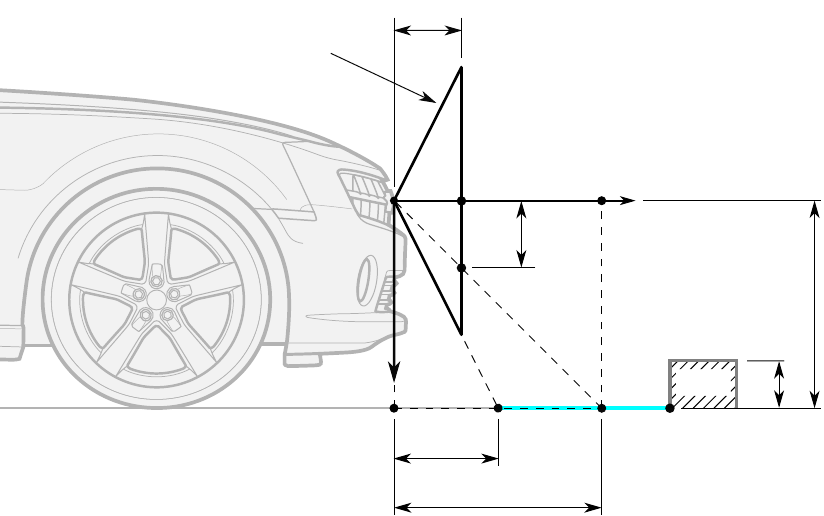
\caption{Employing a monocular camera to measure the distance between a point from the road plane and the camera.}
\label{fig:cameraDistanceMeasurement}
\end{figure}

The plane $y_\mathrm{C} z_\mathrm{C}$ of camera's coordinate frame is illustrated in Fig.~\ref{fig:cameraDistanceMeasurement}.
Let $\mathbf{P} \in \mathbb{R}^3$ is a point from the road plane with coordinates $\mathbf{P} = \begin{bmatrix} x_\mathrm{P} , y_\mathrm{P} , z_\mathrm{P} \end{bmatrix}^\top = \begin{bmatrix}x_\mathrm{P} , H_\mathrm{C} , D_\mathrm{P} \end{bmatrix}^\top$ in camera's coordinate frame.
Its projection onto the image plane is the point $\mathbf{p} = \begin{bmatrix}x_\mathrm{p} , y_\mathrm{p} , 1 \end{bmatrix}^\top \in \mathbb{P}^2$ w.r.t. the frame originating at the principal point $\mathbf{c}$. The relation between their coordinates is
\begin{equation}
  \label{eq:Pproj}
  \mathbf{p} =  \frac{K_\mathrm{c} \mathbf{P}}{D_\mathrm{P}},
\end{equation}
where $K_\mathrm{c} = \begin{bmatrix}
  f_x & 0 & 0 \\ 0 & f_y & 0 \\ 0 & 0 & 1
\end{bmatrix}$, $f_x$ and $f_y$ are camera's focal lengths along its $x_\mathrm{C}$ and $y_\mathrm{C}$ axes, respectively.
From (\ref{eq:Pproj}) we get
  $y_\mathrm{p} = \frac{f_y H_\mathrm{C}}{D_\mathrm{P}}$,
which means that $y_\mathrm{p}$ depends only on $D_\mathrm{P}$, since $f_y$ and $H_\mathrm{C}$ are practically constants.
Furthermore, if we invert the equation, we can calculate the distance $D_\mathrm{P}$ of every point $\mathbf{P}$ from the road plane just by using the vertical coordinate $y_\mathrm{p}$ of its projection $\mathbf{p}$ from the image. Hence, for curb's reference point $\mathbf{P}_\mathrm{U}$ (Fig.~\ref{fig:cameraPos}) can be rewritten
\begin{equation}
  \label{eq:measuringDistances}
  D_\mathrm{U} = \frac{f_y H_\mathrm{C}}{y_\mathrm{U}^{}},
\end{equation}
where $y_{\mathrm{U}}^{}$ is the vertical coordinate of $\mathbf{P}_\mathrm{U}$'s projection in the image.

\begin{figure}[t]
	\centering
	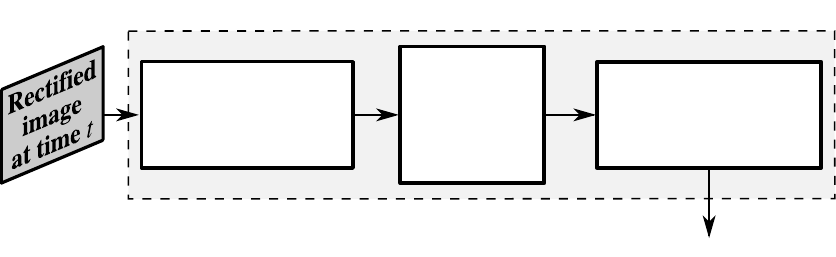
	\caption{Curb candidates detection in a single frame.}
	\label{fig:flowChart-curb_candidates_detection}
\end{figure}

\subsection{Detect curb candidates in an image}
\label{sec:detectingCurbsInImage}

Here we describe our approach for detecting a curb in the image.
The algorithm is inspired by the paradigm for boosted attentional
cascade presented in \cite{ViolaJones}, but instead of using a cascade of
boosted classifiers with gradually increasing complexity, our cascade consists of various filtering techniques.
Thus, the amount of data being processed is greatly reduced by rejecting image regions which do not contain curb features.
As result, only positively classified curb candidates are left for further temporal analysis.

Fig.~\ref{fig:flowChart-curb_candidates_detection} illustrates our
curb detection pipeline, which consists of three consecutive operations:

\subsubsection{Curb scale-invariant edges extraction}
\label{sec:curbEdgeDetection}

\begin{figure}
\centering
\def\svgwidth{\columnwidth}
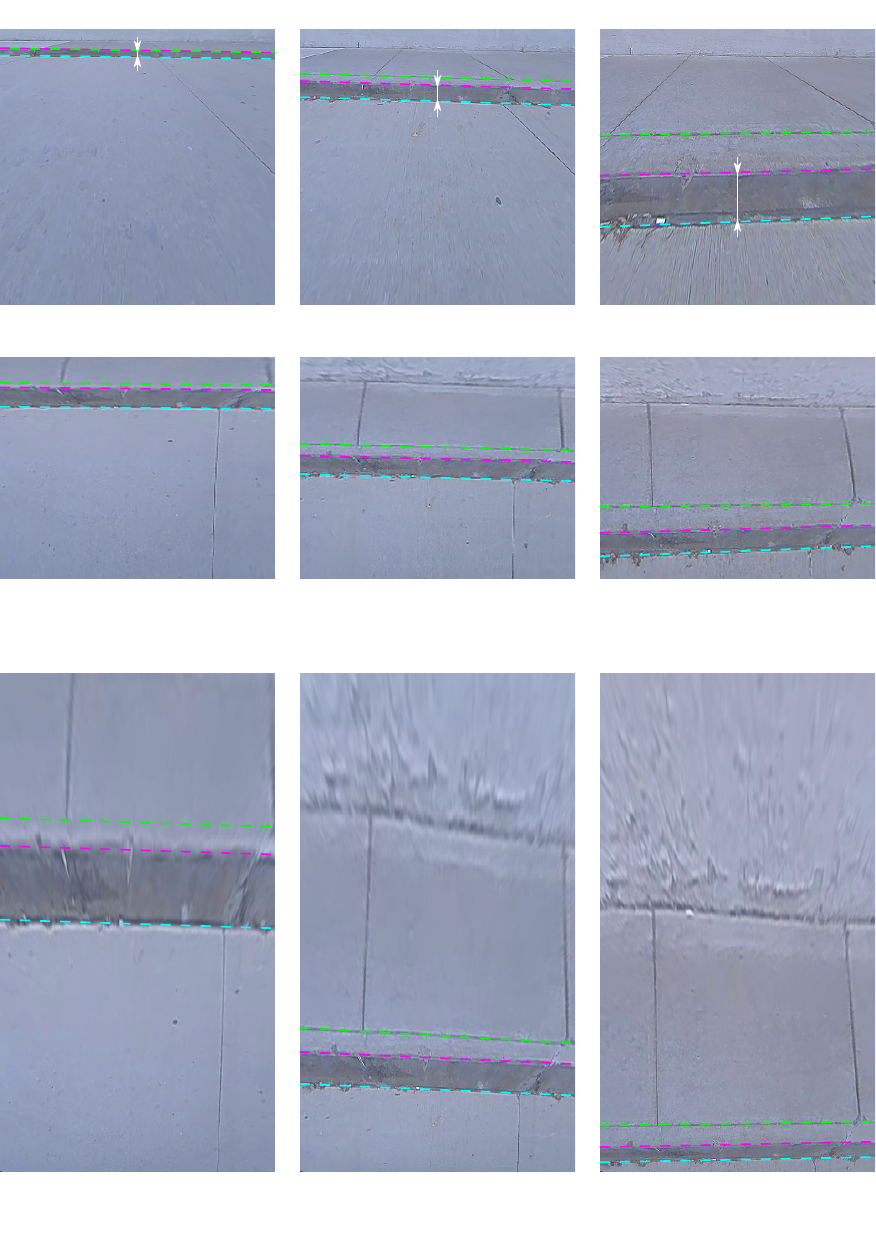
\caption{A set of three curb images taken at three different distances $D_\mathrm{U}$ from the same video sequence.
	They demonstrate the issues related to edge detection and our approach to solve them.}
\label{fig:curbSizeVsDist}
\end{figure}

As we have already mentioned earlier, the primary features which we exploit are the curb's edges (Fig.~\ref{fig:curbDefinitions}).
The straight lines in the image are extracted by the well known combination of
\textit{Hough transform} (HT) \cite{HoughTransform} and the \textit{Canny edge detector} \cite{Canny} applied to the camera images.
Also, as we have already described in Section~\ref{sec:Curb}, the least sufficient condition to perform successful curb detection in the image at time $t$ is that both curb's edges projections $\mathbf{e}_1$ and $\mathbf{e}_2$ are correctly detected.
They define curb's frontal face projection in the image on which we are going to emphasize here.

Let $R_\mathrm{U}$ is curb's frontal face vertical spatial sampling rate in the image
\begin{equation}
  \label{eq:R_U}
  R_\mathrm{U} = \frac{h_\mathrm{U}}{H_\mathrm{U}} = \frac{f_y}{D_\mathrm{U}} \; \left ( \frac{\text{px}}{\text{cm}} \right ),
\end{equation}
where $h_\mathrm{U}$ is the curb's frontal face projection height in the image. Essentially, $R_\mathrm{U}$ provides information about how many sampling locations (pixels) are used by the camera to represent every centimeter of a vertical line from curb's frontal face, which is located at the distance $D_\mathrm{U}$ from the camera.
As a consequence from (\ref{eq:R_U}), $R_\mathrm{U}$ is a non-constant function with respect to $D_\mathrm{U}$, which is demonstrated graphically in Fig.~\ref{fig:curbSizeVsDist}a.
The figure shows three sample images of the same curb taken at 3 different distances $D_\mathrm{U}$.
It is obvious that $h_\mathrm{U}$ (marked in the figure) varies significantly.
In particular, based on our system's setup the ratio
\begin{equation}
  \frac{R_\mathrm{U}(D_\mathrm{U} = D_\text{min})}{R_\mathrm{U}(D_\mathrm{U} = D_\text{max})} = \frac{D_\text{max}}{D_\text{min}} \approx 17.5,
\end{equation} 
i.e.\ the projection of a curb taken at the distance $D_\mathrm{U} = D_\text{min}$ will contain about 17.5 times more pixel information than the projection of the same curb taken at a distance $D_\mathrm{U} = D_\text{max}$.

Our system relies on detecting curb's edges. Principally, edge detection aims in finding the points of discontinuity in the image brightness by incorporating either its first- or second-order derivatives. The \textit{Canny edge detector}, for example, uses first-order operators, such as Sobel \cite{Sobel1968,duda1973pattern}, Scharr \cite{jähne1999handbook}, Prewitt \cite{lipkin1970picture} or Roberts cross \cite{Roberts63} to estimate brightness gradient magnitude and direction.
Having such a significant variation of $R_\mathrm{U}$, though, results in volatile gradient magnitude over curb's edges projections, thus inconsistent curb detection.
In order to overcome this problem, we have derived an image remapping technique --
\textit{Inverse Perspective-compressing Mapping} (IPcM),
which aims to equalize the spatial sampling rate of the curb in the image (Fig.~\ref{fig:curbSizeVsDist}b).
After warping the image, the detection of curb's edges
tends to be much more steady and robust through the entire CDD.
Also, the original trapezoidal shapes of CDD and CSR in the image are transformed into rectangles.
Detailed explanation about the derivation of our technique can be found in Appendix~\ref{sec:curbScaleInvariantRemapping}.

The classical \textit{Inverse perspective mapping} (IPM) normalizes the spatial sampling rate of the road and any other parallel to it plane (Fig.~\ref{fig:curbSizeVsDist}c), but not for the orthogonal frontal curb's face, as can be seen on the figure. Notably, $h_\mathrm{U}$ is still dependent on $D_\mathrm{U}$ and varies considerably. The difference is that after the transformation their relation is proportional.
Moreover, the IPM transform introduces an additional issue, which can be easily acknowledged from the figure.
The output image suffers from gradually increasing interpolation smoothing, mainly along its vertical axis.
It is caused by the irregular density of camera pixels sampling locations in the 3D scene
(the density of the points $\mathbf{P}_i$ in Fig.~\ref{fig:scaleInvariant1} is variable).


Let $\mathcal{L} = \{ \mathbf{l}_i \}_1^{N_\mathcal{L}}$ be the set of straight lines in the
image at time $t$, detected by applying edge detection to the IPcM remapped image and then transforming the lines back to the original image space. $N_\mathcal{L}$ is the size of the set and $\mathbf{l}_i = \begin{bmatrix} a_i, b_i \end{bmatrix}^\top \in \mathbb{R}^2$ are vectors representing the individual lines from the set by their parametric form: $a_i$ -- slope and $b_i$~-- intercept.
This procedure is expected to produce significant amount of outliers. That's exactly what we are aiming at.
The purpose of the following processing blocks of the flow diagram (Fig.~\ref{fig:flowChart-curb_candidates_detection}) is rejection of everything, which is not associated with the curb.
As the curb edges are expected to be among the longest ones in the 
CSR (extending through its full width), we can reduce the processing time by using just the six lines with the highest voting scores from HT ($N_\mathcal{L} \in [0,6]$).
The other types of edges from different geometric shapes, which could possibly be presented in the image (from sidewalk pavement, tiles, etc.), usually have much shorter length, hence lower voting scores.

Finally, we examine the detected lines as a set of points in their parametric space and try to find clusters.
We consider every cluster as a noisy representation of a single edge.
Thus, all the members in a cluster are replaced by its mean.

\subsubsection{Forming curb candidates set}
\label{sec:curbParamsEst}

We construct two new sets~-- $\mathcal{G}_2$ and $\mathcal{G}_3$, which contain all possible combinations of 2- and 3-tuples, respectively,
of non-intersecting lines $\mathbf{l}_i$ from $\mathcal{L}$, i.e.
\begin{equation}
  \begin{split}
    \mathcal{G}_2 = & \left \{ \left ( \mathbf{g}_{21}^{(j)}, \mathbf{g}_{22}^{(j)} \right )_j \right \} = \left \{ \left ( \mathbf{l}_p, \mathbf{l}_q \right )_j \right \} : \\
    & p,q=1 \dots N_\mathcal{L}, \; p \neq q, \; b_p > b_q \\
    & \mathbf{l}_p^{(\mathrm{h})} \times \mathbf{l}_q^{(\mathrm{h})} \notin I_t,\footnotemark \\ & j = 1 \dots N_{\mathcal{G}_2},\; N_{\mathcal{G}_2} \leq \binom{N_\mathcal{L}}{2}
  \end{split}
\end{equation}
and
\footnotetext{Due to the properties of the perspective projection, the relative position and orientation of the camera,
	the road plane and the curb and their shapes,
	it is impossible that any of the lines $\mathbf{e}_{1}$, $\mathbf{e}_{2}$ and $\mathbf{e}_{3}$ have common points in the image $I_t$.
	Therefore, all the 2- and 3-tuples members of $\mathcal{G}_2$ and $\mathcal{G}_3$, respectively, which contain intersecting lines are rejected.}
\begin{equation}
  \begin{split}
    \mathcal{G}_3 = & \left \{ \left ( \mathbf{g}_{31}^{(k)}, \mathbf{g}_{32}^{(k)}, \mathbf{g}_{33}^{(k)} \right )_k \right \} = 
    \left \{ \left ( \mathbf{l}_p, \mathbf{l}_q, \mathbf{l}_r \right )_k \right \} : \\
    & p,q,r=1 \dots N_\mathcal{L}, \; p \neq q \neq r, \; b_{p} > b_{q} > b_{r} \\
    & \mathbf{l}_p^{(\mathrm{h})} \times \mathbf{l}_q^{(\mathrm{h})} \notin I_t,\footnotemark[\thefootnote] \\
    & \mathbf{l}_p^{(\mathrm{h})} \times \mathbf{l}_r^{(\mathrm{h})} \notin I_t,\footnotemark[\thefootnote] \\
    & \mathbf{l}_q^{(\mathrm{h})} \times \mathbf{l}_r^{(\mathrm{h})} \notin I_t,\footnotemark[\thefootnote] \\
    & k = 1 \dots N_{\mathcal{G}_3}, \; N_{\mathcal{G}_3} \leq \binom{N_\mathcal{L}}{3},
  \end{split}
\end{equation}
where $\mathbf{l}_{p,q,r}^{(\mathrm{h})}$ are the homogeneous representations of the lines $\mathbf{l}_{p,q,r}$,
$I_t$ is the camera's image at time $t$ and $N_{\mathcal{G}_{2,3}}$ are the sizes of the two sets.
The lines in every tuple are sorted in ascending order of their vertical placement in the image, i.e.\ in descending order of lines' intercept parameter $b_i$\footnote{Note that the vertical image axis $v$ points downwards.}. 

A tuple of three lines is sufficient to estimate all of the four curb template parameters (\ref{eq:curbTemplateParameters}), rather than the 2-lines case,
where we omit $E_\mathrm{U}$ in favor of estimating the other three more important parameters: $\widehat{D}_\mathrm{U}$, $\widehat{\Theta}_\mathrm{U}$ and $\widehat{H}_\mathrm{U}$.
In order to shorten our presentation we are going to describe the triplet case only, because it is more general, complex and $\mathcal{G}_2$ could be considered as a special case subset of $\mathcal{G}_3$.

The next step is fitting the curb template's projection in the image to each individual triplet in $\mathcal{G}_3$.
Every line from $k$-th triplet is explicitly related to a specific curb's edge, because they will always appear in the image in the same vertical order. Therefore, $\mathbf{g}_{31}^{(k)}$ represents $\mathbf{e}_1$, $\mathbf{g}_{32}^{(k)} \rightarrow \mathbf{e}_2$ and $\mathbf{g}_{33}^{(k)} \rightarrow  \mathbf{e}_3$.
The objective of the fitting is to estimate the optimal values of the parameters $\widehat{\mathbf{x}}_{k} = \begin{bmatrix} \widehat{D}_\mathrm{U}^{(k)}, \widehat{\Theta}_\mathrm{U}^{(k)} , \widehat{H}_\mathrm{U}^{(k)} , \widehat{E}_\mathrm{U}^{(k)} \end{bmatrix} ^\top$, such that curb's template edges projections in the image $\mathbf{e}_{1,2,3}$ are aligned
to their corresponding lines from the triplet $\mathbf{g}_{31,32,33}$ in the best possible way.

\begin{figure}[t]
\centering
\def\svgwidth{\columnwidth}
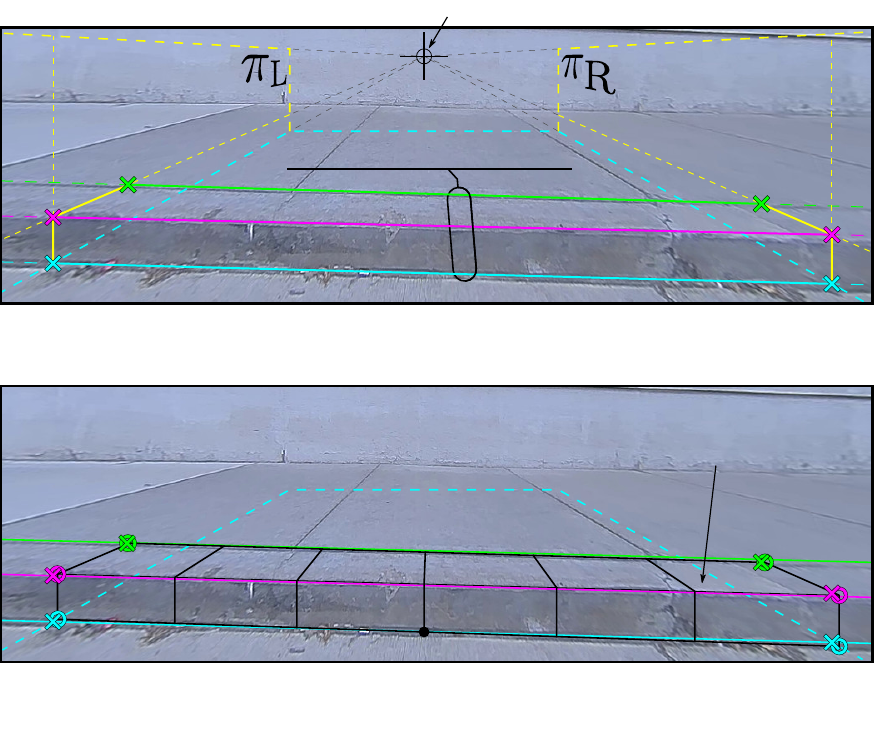
\caption{Curb template fitting. First, the position of the target control points for $k$-th line triplet from $\mathcal{G}_{3}$ (depicted by the crosses) are estimated by following a reasoning based on the principles of the perspective transform and the simplified curb geometry defined in Section~\ref{sec:Curb}.
Second, 3D curb template is fitted to every line tuple by minimizing the distance between its control points (depicted by circles)
and the target control points through adjusting templates parameters $\hat{\mathbf{x}}$.}
\label{fig:curbTemaplteFitting}
\end{figure}

\begin{figure} 
\centering
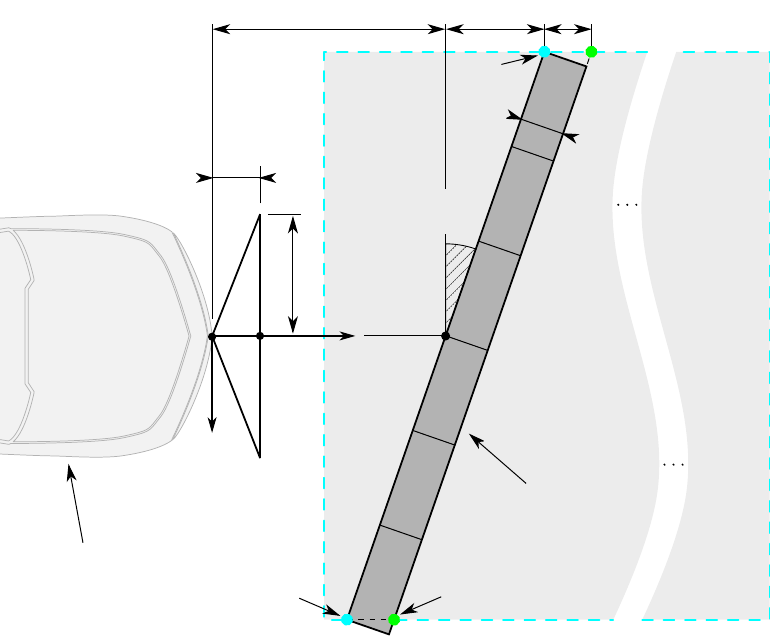
\caption{Top view of the vehicle, camera and curb template, showing the location of the control points $\widehat{\mathbf{P}}_m : m = 1,\dots,6$.}
\label{fig:curbTemplatePointProjectionsCalc}
\end{figure}

We evaluate the similarity of two $\mathbb{R}^2$ lines by calculating the Euclidean distance between two pairs of corresponding points laying on each of them, which 
we call \textit{control points}. Hence, to fit the template to a triplet of lines we need to use six pairs of control points.
In order to define their position, we need to introduce the planes $\mathbf{\Pi}_\mathrm{L}$ and $\mathbf{\Pi}_\mathrm{R}$ and their projections in the image $\mathbf{\pi}_\mathrm{L}$ and $\mathbf{\pi}_\mathrm{R}$ (Fig.~\ref{fig:curbTemaplteFitting}a).
They intersect $\mathbf{\Pi}_\mathrm{G}$ in $\mathbf{B}_\mathrm{L}$ and $\mathbf{B}_\mathrm{R}$ and $\mathbf{\Pi}_\mathrm{L,R} \perp \mathbf{\Pi}_\mathrm{G}$.
Now curb template's control points are defined by intersecting its edge lines $\widehat{\mathbf{E}}_{1,2,3}$ with $\mathbf{\Pi}_\mathrm{L}$ and $\mathbf{\Pi}_\mathrm{R}$,
which results in two $\mathbb{R}^3$ point triplets: $(\widehat{\mathbf{P}}_{1}, \widehat{\mathbf{P}}_{2}, \widehat{\mathbf{P}}_{3})$ and $(\widehat{\mathbf{P}}_{4}, \widehat{\mathbf{P}}_{5}, \widehat{\mathbf{P}}_{6})$ -- Fig.~\ref{fig:curbTemplatePointProjectionsCalc}.
The calculation of their coordinates in camera's coordinate frame is significantly simplified due to systems geometrical setup, namely
\begin{equation}
\begin{aligned}
  \widehat{\mathbf{P}}_{1} & = [ & -W_{\text{max}}&, & H_\mathrm{C}&,                          & \widehat{D}_\mathrm{U}& + \Delta D_1              & ]^\top \\
  \widehat{\mathbf{P}}_{2} & = [ & -W_{\text{max}}&, & H_\mathrm{C}& - \widehat{H}_\mathrm{U}, & \widehat{D}_\mathrm{U}& + \Delta D_1              & ]^\top \\
  \widehat{\mathbf{P}}_{3} & = [ & -W_{\text{max}}&, & H_\mathrm{C}& - \widehat{H}_\mathrm{U}, & \widehat{D}_\mathrm{U}& + \Delta D_1 + \Delta D_2 & ]^\top \\
  \widehat{\mathbf{P}}_{4} & = [ &  W_{\text{max}}&, & H_\mathrm{C}&,                          & \widehat{D}_\mathrm{U}& - \Delta D_1              & ]^\top \\
  \widehat{\mathbf{P}}_{5} & = [ &  W_{\text{max}}&, & H_\mathrm{C}& - \widehat{H}_\mathrm{U}, & \widehat{D}_\mathrm{U}& - \Delta D_1              & ]^\top \\
  \widehat{\mathbf{P}}_{6} & = [ &  W_{\text{max}}&, & H_\mathrm{C}& - \widehat{H}_\mathrm{U}, & \widehat{D}_\mathrm{U}& - \Delta D_1 + \Delta D_2 & ]^\top
\end{aligned},
\end{equation}
where
\begin{equation}
  \begin{split}
      \Delta D_1 & = W_\text{max} \tan \widehat{\Theta}_\mathrm{U} \\
      \Delta D_2 & = \frac{\widehat{E}_\mathrm{U}}{\cos \widehat{\Theta}_\mathrm{U}}
  \end{split}.
\end{equation}
Afterwards, we can estimate their projections in the image as follows
\begin{equation}
  \widehat{\mathbf{p}}_{m}(\widehat{\mathbf{x}}) = K \widehat{\mathbf{P}}_{m},
\end{equation}
where $m=1 \dots 6$ and $K = \begin{bmatrix}
  f_x & 0 & c_x \\ 0 & f_y & c_y \\ 0 & 0 & 1
\end{bmatrix}$ is the camera matrix, $c_x$ and $c_y$ are camera's principal point $\mathbf{c}$ coordinates along the horizontal and vertical axes, respectively. It should be noted that since $K$, $w_\text{max}$ and $H_\mathrm{C}$ are constants, $\widehat{\mathbf{p}}_{m}$ is function only of the curb parameters vector $\widehat{\mathbf{x}}$.

The estimation of $k$-th triplet control points\footnote{We will call them \textit{target control points}.} positions in the image is demonstrated in Fig.~\ref{fig:curbTemaplteFitting}a. It follows the presumptions of template's control points location in the image and incorporating the properties of perspective projection.
Firstly, we intersect $\mathbf{g}_{31}^{(k)}$ with $\mathbf{b}_\mathrm{L,R}^{}$ and thus estimate the control points $\mathbf{p}_{31}^{(k)}$ and 
$\mathbf{p}_{34}^{(k)}$ (depicted with cyan crosses in the figure).
\begin{equation}
  \mathbf{p}_{31}^{(k)} = \mathbf{b}_{\mathrm{L}}^{} \times \mathbf{g}_{31}^{(k)} \; \; \mathrm{and} \; \;
  \mathbf{p}_{34}^{(k)} = \mathbf{b}_{\mathrm{R}}^{} \times \mathbf{g}_{31}^{(k)} .
\end{equation}
We know that curb's front face is vertical, i.e.\ parallel to camera's image plane.
Therefore, we find $\mathbf{p}_{32}^{(k)}$ and $\mathbf{p}_{35}^{(k)}$ by intersecting $\mathbf{g}_{32}^{(k)}$ with the two vertical lines $\mathbf{l}_{\mathrm{Lv}}^{(k)}$ and $\mathbf{l}_{\mathrm{Rv}}^{(k)}$ from the figure which pass through $\mathbf{p}_{31}^{(k)}$ and $\mathbf{p}_{34}^{(k)}$, i.e.
\begin{equation}
  \mathbf{p}_{32}^{(k)} = \mathbf{l}_{\mathrm{Lv}}^{(k)} \times \mathbf{g}_{32}^{(k)} \; \; \mathrm{and} \; \;
  \mathbf{p}_{35}^{(k)} = \mathbf{l}_{\mathrm{Rv}}^{(k)} \times \mathbf{g}_{32}^{(k)} .
\end{equation}
In Fig.~\ref{fig:curbTemaplteFitting} these control points are depicted by magenta crosses.

Camera's principal point $\mathbf{c}$ is the vanishing point (center of perspective), where the projections in the image of all $\mathbb{R}^3$ lines parallel to $z_\mathrm{C}$ converge. 
Hence, the lines $\mathbf{l}_\mathrm{Lh}^{(k)}$ and $\mathbf{l}_\mathrm{Rh}^{(k)}$ in Fig.~\ref{fig:curbTemaplteFitting}a that 
constitute the projections of the intersections of curb's top face and the planes $\mathbf{\Pi}_\mathrm{L,R}$ will pass through it and we can estimate
the positions of the last two target control points $\mathbf{p}_{33}^{(k)}$ and $\mathbf{p}_{36}^{(k)}$ as follows
\begin{equation}
    \mathbf{l}_\mathrm{Lh}^{(k)} = \mathbf{c} \times \mathbf{p}_{32}^{(k)} \; \; \mathrm{and} \; \;
    \mathbf{l}_\mathrm{Rh}^{(k)} = \mathbf{c} \times \mathbf{p}_{35}^{(k)},
\end{equation}
\begin{equation}
  \mathbf{p}_{33}^{(k)} = \mathbf{l}_\mathrm{Lh}^{(k)} \times \mathbf{g}_{33}^{(k)} \; \; \mathrm{and} \; \;
  \mathbf{p}_{36}^{(k)} = \mathbf{l}_\mathrm{Rh}^{(k)} \times \mathbf{g}_{33}^{(k)}.
\end{equation}
They are depicted by the green crosses in the figure.

After we have derived the equations of all control points in the image, we can define the objective function, which we are going to optimize in order to fit the curb template's projection in the image to the lines of $k$-th triplet from $\mathcal{G}_3$.
First, let the distance $r_{3m}^{(k)}$ between two corresponding control points in the image be the $L^2$-norm for their difference, i.e.
\begin{equation}
  r_{3m}^{(k)}(\widehat{\mathbf{x}}) = \left \| \mathbf{p}_{3m}^{(k)} - \widehat{\mathbf{p}}_{3m}^{}(\widehat{\mathbf{x}}) \right \| _2 ,
\end{equation}
where $m=1\dots6$. Then, we construct the error vector
\begin{equation}
  \mathbf{r}_{3}^{(k)}(\widehat{\mathbf{x}}) = \begin{bmatrix} r_{31}^{(k)} , r_{32}^{(k)} , \dots , r_{36}^{(k)} \end{bmatrix}^\top
\end{equation}
and the curb template's parameters that produce the best fit to the $k$-th triplet are determined as follows 
\begin{equation}
  \begin{split}
  \widehat{\mathbf{x}}_{3}^{(k)} & = \argmin_{\widehat{\mathbf{x}}} \alpha_{D} \left \| \mathbf{r}_{3}^{(k)}(\widehat{\mathbf{x}}) \right \|_2 \\
   & \equiv \argmin_{\widehat{\mathbf{x}}} \alpha_{D} \sum_{m=1}^{6} \left [ r_{3m}^{(k)}(\widehat{\mathbf{x}}) \right ]^2
  \end{split},
\end{equation}
where $\alpha_{D} = \frac{\widehat{D}_{\mathrm{U}}}{D_\text{max}}$ is a normalization term, which regularizes the dependence between
the template's re-projection error in the image and the distance $\widehat{D}_\mathrm{U}$.
As the minimization of $L^2$-norm of a vector is equivalent to minimizing the sum of its squared elements and we have closed form differentiable solution for $\mathbf{r}_{3}^{(k)}$, we can incorporate Levenberg-Marquardt optimization algorithm \cite{LMA1, LMA2}.
Fig.~\ref{fig:curbTemaplteFitting} illustrates an example of a successfully optimized (fitted) curb template.

After fitting the curb template to all the line tuples in $\mathcal{G}_2$ and $\mathcal{G}_3$, we build 
the curb candidates set
\begin{equation}
  \widehat{\mathcal{X}} = \left \{ \widehat{\mathbf{x}}_{2}^{(1)}, \widehat{\mathbf{x}}_{2}^{(2)}, \dots, \widehat{\mathbf{x}}_{2}^{(N_{\mathcal{G}_2})}, \widehat{\mathbf{x}}_{3}^{(1)}, \widehat{\mathbf{x}}_{3}^{(2)}, \dots, \widehat{\mathbf{x}}_{3}^{(N_{\mathcal{G}_3})} \right \}.
\end{equation}
The next task is rejecting the outliers from $\widehat{\mathcal{X}}$. The first level of filtering is described in the next section, where curb candidates are rejected based on their appearance in the image at time $t$.
Afterwards, among the ``survivals" only the ones, whose parameters follow the prediction,
based on temporal analysis of the previous frames, are selected. This procedure is described in Section~\ref{sec:curbTracking}.

\subsubsection{Appearance-based curb candidates filtering}
\label{sec:curbCandidatesFiltering}

\begin{figure}[t]
	\centering
	\def\svgwidth{0.8\columnwidth}
	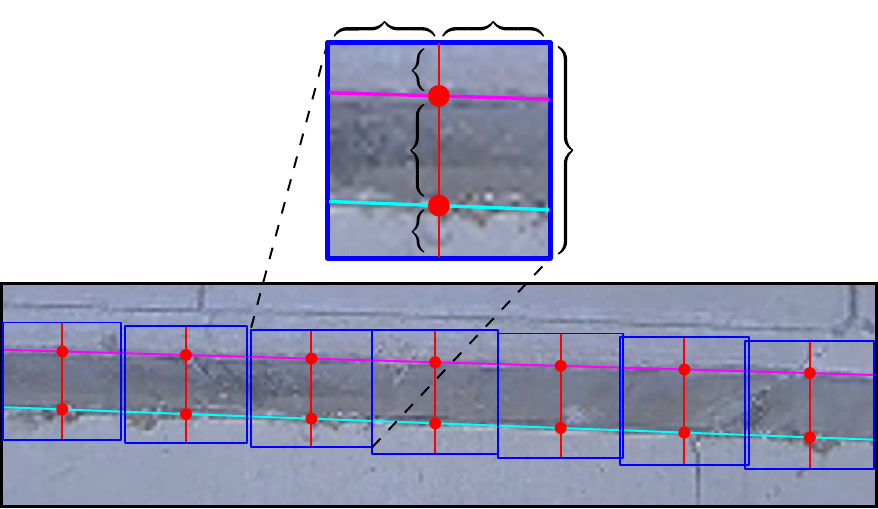
	\caption{Sampling square uniformly distributed windows along the frontal face projection of a curb template in the IPcM image to extract HOG features vector.}
	\label{fig:SVM_subwindows}
\end{figure}

It is very unlikely that a trustworthy curb detection could be accomplished by employing only the straight curb's edges from the image, since they are not informative enough.
I.e. relying only on the curb's geometry won't bring down the entropy to levels that the algorithm can reliably discriminate between curb and non-curb shapes.
At this stage we try to reject the outliers in $\widehat{\mathcal{X}}$ by exploiting curb's appearance in the image. Thus, we have built an object detector
based on a \textit{Support Vector Machine} (SVM) and \textit{Histograms of Oriented Gradients} (HOG) features \cite{HOG}.

\begin{figure}[t]
\centering
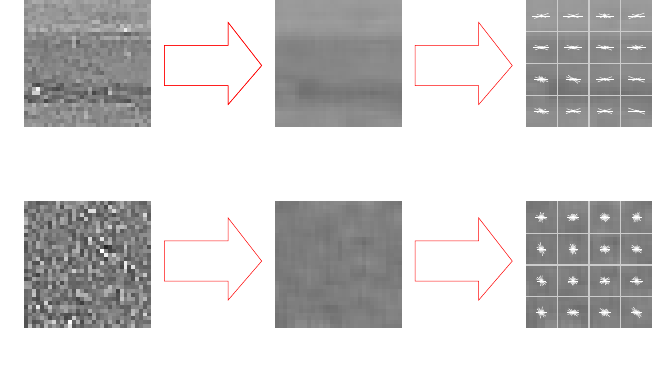
\caption{Classification samples and their HOG representations.}
\label{fig:HOG-Samples}
\end{figure}

Curbs occupy areas in the image, which have the shape of thin, mostly horizontal, stripes.
They are characterized by a couple of distinctive transitions of pixels' brightness along the vertical axis,
caused by the differences in reflecting/scattering properties of the individual curb's surfaces.
In the case of curb detection in images the HOG is a suitable descriptor, becuase it accounts the direction
of that transitions and a machine learning algorithm can be trained to discriminate among curb and non-curb image patterns.
Moreover, HOG is also popular with its computational efficiency.

We extract the HOG features from the IPcM image, because the vertical size of the curb is invariant to $D_\textrm{U}$.
Fig.~\ref{fig:SVM_subwindows} illustrates our sampling approach. Seven uniformly distributed square windows
are sampled along curb template frontal face.
Only the two frontal curb edges are needed to accomplish that operation. 
The size of the patches is determined by the distance between the two edge lines in vertical direction (on the figure, depicted by $a$).
Each window is scaled down to a fixed size of \mbox{$32 \times 32$}~px -- Fig.~\ref{fig:HOG-Samples}.
In order to eliminate the influence of the high-frequency components, we apply a smoothing Gaussian filter to the windows before calculating their HOG features. In the end, the pixel information of each of them is converted to 288-dimensional HOG feature vector, which is fed to the pre-trained linear SVM classifier \cite{liblinear}.
Fig.~\ref{fig:HOG-Samples} illustrates examples of a positive (a) and a negative (b) windows. Even an unexperienced human eye can easily notice the considerable difference between the gradient histograms of the two samples.

We define our classification problem in a \textit{Multiple Instance Learning} manner.
The set of the seven feature vectors sampled form the same curb template from $\widehat{\mathcal{X}}$ form a bag of instances $\mathcal{B}$.
We define two types of bags -- positive ($\mathcal{B}^+$) and negative ($\mathcal{B}^-$).
If the majority of the instances in a bags are positive, then that bag is considered positive.
And respectively,
if the majority of the instances are negative, then the bag is also negative.
Thus, our classifier will be more robust against outliers in the bags.
In other words, instances, which represent a curb,
but are sampled from a regions which contain vertical cracks or joints, is expected to be placed
far from the other instances of the same class in the feature space.
The curb candidates from $\widehat{\mathcal{X}}$ that are approved by this classification procedure form the final in-frame curb candidates set
\begin{equation}
  \widetilde{\mathcal{X}} = \left \{ \widetilde{ \mathbf{x}}_1, \widetilde{\mathbf{x}}_2, \dots, \widetilde{\mathbf{x}}_{N} \right \},
\end{equation}
where $N \leq N_{\mathcal{G}_2} + N_{\mathcal{G}_3}$ is its size. Note that if $\widetilde{\mathcal{X}} = \varnothing$,
the curb detection is unsuccessful and thus, we accept that such frames does not contain curbs.

\subsection{Tracking the curb through the time}
\label{sec:curbTracking}

\begin{figure}[t]
\centering
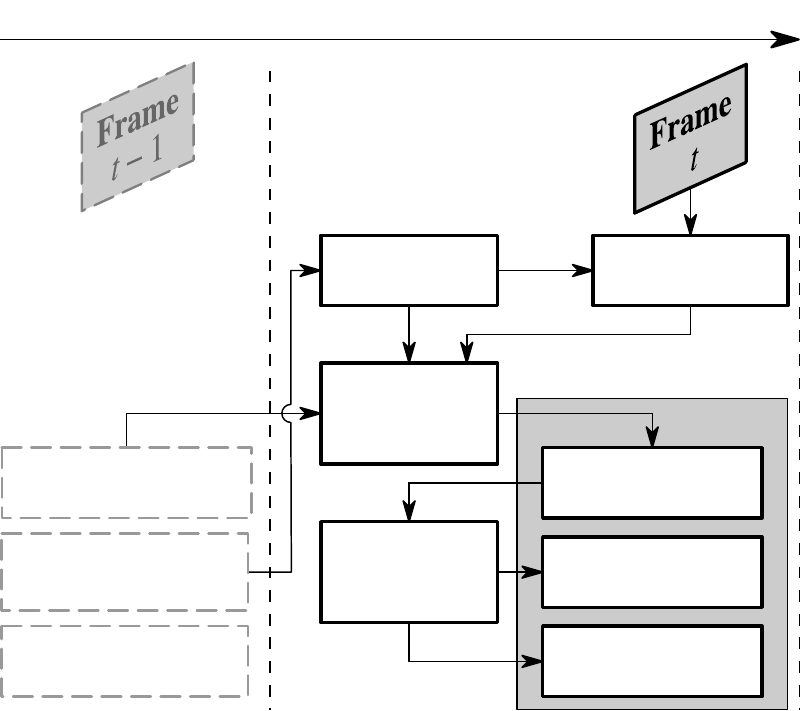
\caption{Flow diagram of the \textit{``Curb tracking''} mode.}
\label{fig:flowChart}
\end{figure}

Here we present our scheme for curb tracking in the time domain at frame-to-frame basis.
The reasoning here is based on the assumption that curb false candidates
are result of faulty lines detection that occur,
because of the noisy output from the Canny operator and false positive classification by the HOG+SVM.
To the first reason contributes the significant local contrast in the small details produced by the HDR camera we use.
Therefore, we can assume that those faulty detections don't follow a predictable and smooth pattern in the 
time domain.

We have prior knowledge regarding the nature of curb parameters evolution (\ref{eq:curbParams}).
Namely, we know that $H_\mathrm{U}$ and $E_\mathrm{U}$ are constants and $D_\mathrm{U}$ and $\Theta_\mathrm{U}$
evolve smoothly over time, since the vehicle is a physical object whose motion is continuous function of time.
Thus, we model them as autoregressive processes.

The temporal filtering we apply has two alternating modes:
\begin{itemize}
  \item \textit{Collecting initial prediction set}
  \item \textit{Curb tracking} 
\end{itemize}
During the first one, the successful curb candidates set $\widetilde{\mathcal{X}}_{t}$ of the current frame at time $t$
is appended to a finite length buffer $\mathcal{C}$.
When the critical minimum for tracking is reached (5 consecutive successful frames), 
curb parameters prediction lines are estimated and then the mode is switched to \textit{``Curb tracking"}.
I.e.
$\mathcal{C} = \left \{ \widetilde{\mathcal{X}}_{t}, \widetilde{\mathcal{X}}_{t-1} \dots \widetilde{\mathcal{X}}_{t-4} \right \}$,
where $\widetilde{\mathcal{X}}_{t-n} \neq \varnothing: n = 0,\dots,4$.

The prediction lines are used to predict the future system state and to smooth the measurements of the current state,
by taking into account the previous system states.
We assume that within a short span of time (for example 7 frames) the evolution curves of the curb parameters have mainly linear character.
Each curb parameter has its individual prediction line -- $\left ( \mathbf{l}_{D}, \mathbf{l}_{\Theta}, \mathbf{l}_{H}, \mathbf{l}_{E} \right )_{t}$.
Their parameters are estimated by linear regression applied to the corresponding elements of the curb candidate vectors $\widetilde{\mathbf{x}}$
of every possible combination of 5 candidates $\left ( \widetilde{\mathbf{x}}_i^{(t)}, \widetilde{\mathbf{x}}_j^{(t-1)}, \widetilde{\mathbf{x}}_k^{(t-2)}, \widetilde{\mathbf{x}}_l^{(t-3)}, \widetilde{\mathbf{x}}_m^{(t-4)} \right ) : \widetilde{\mathbf{x}}_i^{(t-n)} \in \widetilde{\mathcal{X}}_{t-n},\, n = 0,\dots,4,\, i = 1,\dots,N^{(t)},\, j=1,\dots,N^{(t-1)},\, k=1,\dots,N^{(t-2)},\, l=1,\dots,N^{(t-3)},\, m=1,\dots,N^{(t-4)}$, sampled from $\mathcal{C}$.
The combination, which has minimal fitting error, is chosen to be the prediction samples set $\mathcal{P}_t$ at time $t$.


Now, let's assume that system processing mode at the current time $t$ is \textit{``Curb tracking"}.
The flow chart diagram from Fig.~\ref{fig:flowChart}, presents it graphically.
The first operation is predicting system state at time $t$
\begin{equation}
	\label{eq:curbParametersPrediction}
	\mathbf{\accentset{*}{x}}_t = \begin{bmatrix} \accentset{*}{D}_\mathrm{U}^{(t)}, \accentset{*}{\Theta}_\mathrm{U}^{(t)}, \accentset{*}{H}_\mathrm{U}^{(t)}, \accentset{*}{E}_\mathrm{U}^{(t)} \end{bmatrix}^\top,
\end{equation}
based only on the information
from the previous frames -- $\left ( \mathbf{l}_{D}, \mathbf{l}_{\Theta}, \mathbf{l}_{H}, \mathbf{l}_{E} \right )_{t-1}$.
This gives us information for the approximate position of the curb in the current frame
and length and position of CSR can be updated accordingly before detecting the curb candidates in the second operation.

Then the current frame's prediction set $\mathcal{P}_{t}$ is obtained by 
selecting the closest to the prediction $\mathbf{\accentset{*}{x}}_t$ curb candidate from $\widetilde{\mathcal{X}}_{t}$
and appending it to $\mathcal{P}_{t-1}$.
The last step is estimating the prediction lines set for the current frame 
$\left ( \mathbf{l}_{D}, \mathbf{l}_{\Theta}, \mathbf{l}_{H}, \mathbf{l}_{E} \right )_{t}$ and
the smoothed (filtered) version of the curb state $\bar{\mathbf{x}}_t$, which supposedly contains much less noise than the individual
measurements.

\section{Experimental Results}
\label{sec:experimentalResults}

\subsection{Video dataset} 

We have collected a dataset, which consists of 11 videos captured with a monocular forward-view fisheye HDR camera in typical forward perpendicular parking situations during the bright part of the day in a natural lighting environment~-- Table~\ref{tab:datasetInfo}.
All but one videos contain a single sidewalk curb.
Only video sequence 8 has two perpendicular curbs presented in the CCD.
Two distinct weather conditions were presented at the time of data collection -- clear/sunny, which is characterized with sharp deep shadows and bright highlights creating unreal edges in the image, and shadow (overcast), which is characterized with soft shadows that smoothly grade to highlights.
In sequences 2, 3, 8, 11 the front bottom curb edge is fully or partially obstructed by tree leaves and different kind of debris.
Camera frame rate is approximately 21 fps and the original resolution is $1920\times1080$~pixels.

The distance $D_\mathrm{U}$ \textit{ground truth} (GT) data
is collected by the means of a point-wise LIDAR sensor.
The height $H_\mathrm{U}$ and the depth $E_\mathrm{U}$ of the curb are measured manually with a ruler.
No direct ground truth measurements are made of curb's rotation angle $\Theta_\mathrm{U}$.
It has been implicitly estimated through manually fitted template to the curb appearances in the video frames.
The maximum labeling distance is 5~m.

\begin{table}[t]
	\centering
	\caption{\label{tab:datasetInfo}Dataset details}
	\begin{threeparttable}
	\begin{tabular}{|>{\centering\arraybackslash} m{5mm}|>{\centering\arraybackslash} m{8mm}|>{\centering\arraybackslash} m{8mm}|>{\centering\arraybackslash} m{13mm}|>{\centering\arraybackslash} m{13mm}|>{\centering\arraybackslash} m{9mm}|}
		\hline
		\textbf{Vid. seq. \#} & \textbf{Curb height (cm)} & \textbf{Curb depth (cm)} & \textbf{Weather conditions} & \textbf{Curb/road physical properties} & \textbf{Frames count}\tnote{$\ddagger$}  \\
		\hline
		\hline
		 1 & 11.1 & 20.6 & Clear & Co./As.\tnote{$*$} & 702/374 \\
		\hline
		 2 & 13.3 & 20.6 & Clear & Co./As.\tnote{$*$} & 665/344 \\
		\hline
		 3 & 10.6 & 20.6 & Clear & Co./As.\tnote{$*$} & 626/378 \\
		\hline
		 4 & 16.2 & 15.9 & Shadow & Co./As.\tnote{$*$} & 519/332 \\
		\hline
		 5 & 14.6 & 16.4 & Shadow & Co./As.\tnote{$*$} & 497/321 \\
		\hline
		 6 & 10.5 & 20.6 & Clear & Co./As.\tnote{$*$} & 580/345 \\
		\hline
		 7 & 10.8 & 20.3 & Shadow & Co./As.\tnote{$*$} & 545/318 \\
		\hline
		 8 &  9.8 & 21.6 & Shadow & Co./As.\tnote{$*$} & 521/341 \\
		\hline
		 9 & 11.4 & 20.8 & Shadow & Pa./St.\tnote{$\dagger$} & 486/291 \\
		\hline
		10 &  9.8 & 20.3 & Shadow & Pa./St.\tnote{$\dagger$} & 412/308 \\
		\hline
		11 & 13.7 & 20.8 & Clear & Co./As.\tnote{$*$} & 555/360 \\
		\hline
	\end{tabular}
	\begin{tablenotes}\footnotesize
		\item[$*$] Concrete/Asphalt
		\item[$\dagger$] Painted/Strained
		\item[$\ddagger$] Total number of frames in the sequence/Number of the frames with curb presented in the CCD
	\end{tablenotes}
	\end{threeparttable}
\end{table}

\subsection{Performance}

\begin{figure}
	\centering
	\def\svgwidth{\columnwidth}
	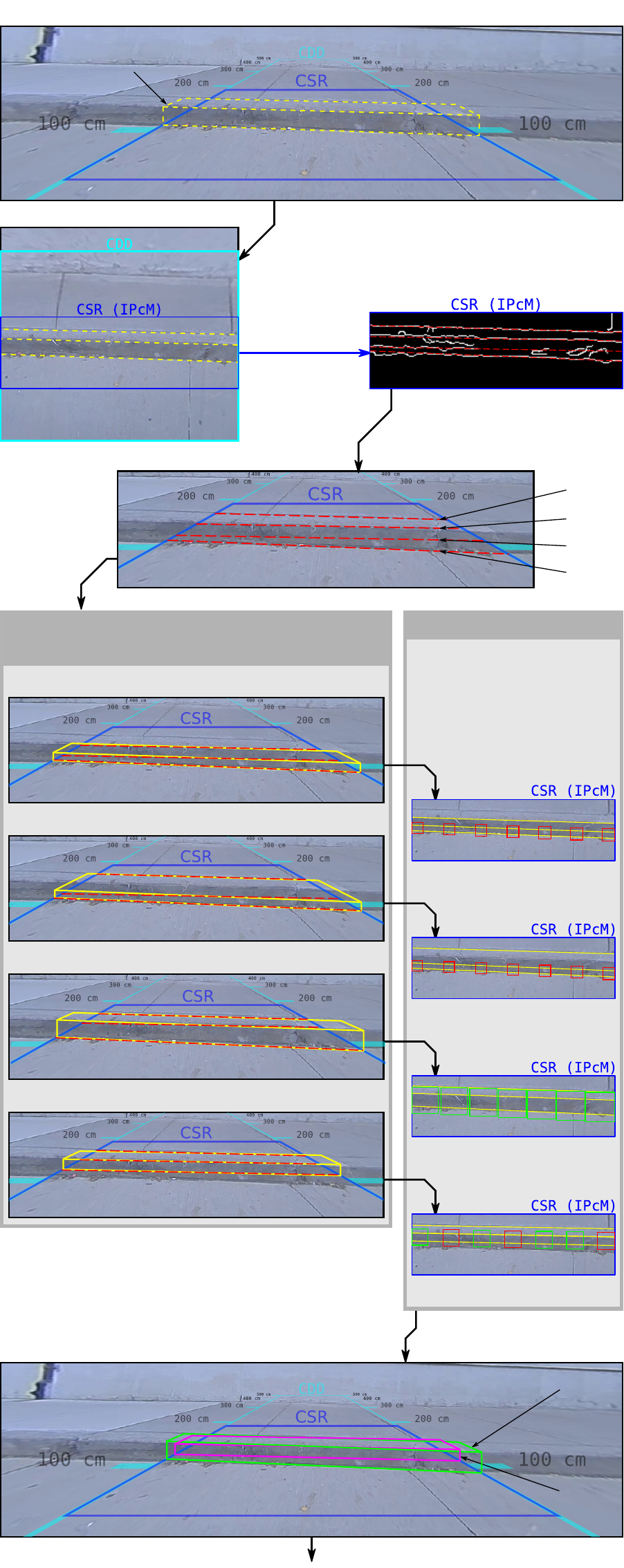
	\caption{Demonstration of system's spatial domain processing procedure for detecting 3-edges curb candidates in frame \#223 from video sequence 4 of our dataset.}
	\label{fig:demo-in-image}
\end{figure}

\begin{figure}
	\centering
	\def\svgwidth{\columnwidth}
	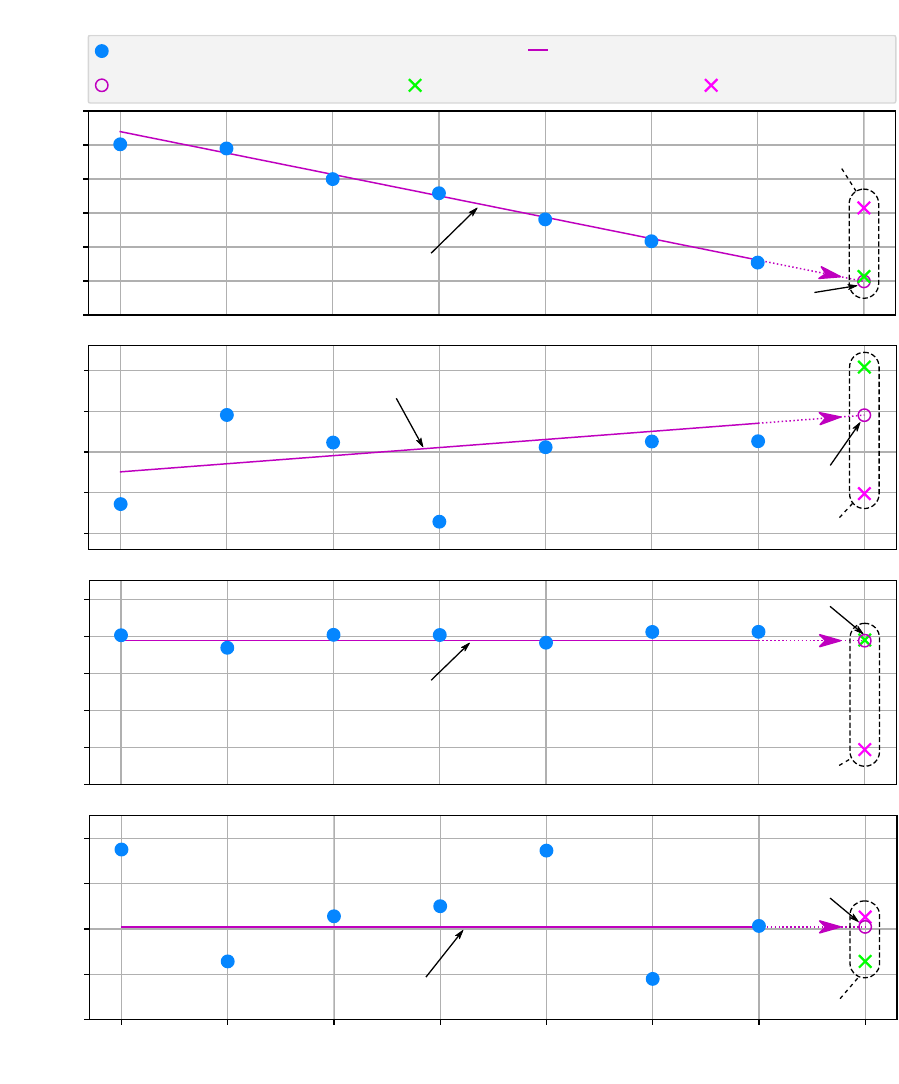
	\caption{Demonstration of system's temporal domain processing procedure for detecting 3-edges curb candidates in frame 223 of video sequence 4.}
	\label{fig:demo-temporal}
\end{figure}

Our curb detector has been implemented as a Python-language program. 
Most of the image processing procedures are accomplished by open source libraries, such as \textit{OpenCV} \cite{opencv_library} and \textit{LibLinear} \cite{liblinear}.
The overall processing performance is evaluated at average $11-12$ fps for Full~HD images and by considering the fact that no  graphical processor (GPU) optimization is used.
In other words, there are opportunities for further significant improvements of the execution speed.

Figures \ref{fig:demo-in-image} and \ref{fig:demo-temporal} present a visual example of our system's processing procedures in real situation.
More specifically, the time of execution $t$ equals to frame number 223 of video sequence 4 from our dataset.
The vehicle is approaching a sidewalk curb and the system has already collected sufficient temporal data
and the current working mode is \textit{``Curb tracking''}.
Both figures correspond to the intra-frame (spatial) and inter-frame (temporal)
curb detection, described in Sections~\ref{sec:detectingCurbsInImage} and \ref{sec:curbTracking}, respectively.

As described in Fig.~\ref{fig:flowChart}, first of all the system predicts
curb parameters $\mathbf{\accentset{*}{x}}_t$ at time $t$.
That process is visualized in Fig.~\ref{fig:demo-temporal},
where the prediction lines from frame $t-1$ are used to
extrapolate all elements of the vector $\mathbf{\accentset{*}{x}}_t$ (\ref{eq:curbParametersPrediction}) independently.
Knowing the approximate location of the curb, the system determines CSR's size and position,
such that curb's frontal face is centered in CSR IPcM remapped region (Fig.~\ref{fig:demo-in-image}a and b).
The next step is extracting the straight lines from it (Fig.~\ref{fig:demo-in-image}c)
supposing that most of them are going to represent curb edges.
The lines are inversely transformed from IPcM remapped image to original space and 
the lines set $\mathcal{L}_t$ is constructed (Fig.~\ref{fig:demo-in-image}d).
As can be observed in the figure, the system has detected 4 lines, 3 of which represent
the three curb edges -- $\mathbf{l}_1$, $\mathbf{l}_3$ and $\mathbf{l}_4$.
The line $\mathbf{l}_2$ is an outlier, which is going to be rejected during
the following processing stages.
Next, the set $\mathcal{G}_3^{(t)}$ is constructed from all possible combinations
of $\mathcal{L}_t$ line triplets\footnote{
	Current example is related to 3-edges curb candidates detection for brevity.
	The procedure for 2-edges ones ($\mathcal{G}_2$) is similar and simpler,
	as curb candidates depth estimations are skipped.
}.
In the current example they are four (Fig.~\ref{fig:demo-in-image}e).
Into each of them 3D curb templates are fitted and their state vectors $\widehat{\mathbf{x}}_3^{(i)}$, $i=1 \dots 4$
constitute 3-lines curb candidates set $\widehat{\mathcal{X}}_t$.
Three of these four candidates are false -- caused by $\mathbf{l}_2$.
In the following validation procedure the seven uniformly distributed square windows are sampled
along each template's frontal face. HOG features are extracted from them and
grouped into bags: $\mathcal{B}_1^{(t)}$, $\mathcal{B}_2^{(t)}$, $\mathcal{B}_3^{(t)}$, $\mathcal{B}_4^{(t)}$.
Each bag contains the HOG feature vectors sampled from the same template.
In Fig.~\ref{fig:demo-in-image}f with green and red colors are depicted the positive and negative labels,
respectively, assigned by the classifier.
From the figure, it is obvious that bags $\mathcal{B}_1^{(t)}$, $\mathcal{B}_2^{(t)}$ and $\mathcal{B}_3^{(t)}$
have very convincing labels, whereas the classifier is ``uncertain'' about $\mathcal{B}_4^{(t)}$ and
the assigned label is falsely positive, since the majority if bags members have positive labels.
At the end of this procedure, the in-frame curb candidates set $\widetilde{\mathcal{X}}_t$ consists of two members
(Fig.~\ref{fig:demo-in-image}g), which are going to be a subject to temporal filtering~-- Fig.~\ref{fig:demo-temporal}.

Among all curb candidates in $\widetilde{\mathcal{X}}_t$, the one closest ot the prediction $\mathbf{\accentset{*}{x}}_t$ is chosen.
As curb parameters are heterogeneous\footnote{Their nature, magnitude and range are different.},
the system compares each one independently, in application-wise importance ascending order\footnote{
	The main purpose of the current system is to avoid collisions between the vehicle and obstacles, such as curbs. Therefore, we can assign figurative importance level to each of curb parameters, from application point of view.
	We consider $D_U$ as the most important parameter, because if the curb is far enough, we know that a collision won't occur, regardless the values of the other curb parameters. If we cannot rely on $D_U$ to infer whether or not the vehicle will collide, we should take into account $H_U$. Therefore, it is considered as the second most important parameter, etc.}.
In Fig.~\ref{fig:demo-temporal}
we can see that $\widehat{\mathbf{x}}_3^{(3)}$ is much closer to $\accentset{*}{D}_\mathrm{U}^{(t)}$,
than $\widehat{\mathbf{x}}_3^{(4)}$ is. Moreover, $\widehat{\mathbf{x}}_3^{(4)}$ is
undeniable outlier, since the distance to $\accentset{*}{D}_\mathrm{U}^{(t)}$ is way longer than
the width of a 99\% confidence interval,
which proves that it has not been drawn from the same distribution as the prediction samples
in $\mathcal{P}_{t-1}$ and $\widehat{\mathbf{x}}_3^{(3)}$. Even if we propagate the temporal analysis
deeper to the next most important parameter~-- $H_\text{U}$,
we can clearly see the $\widehat{\mathbf{x}}_3^{(3)}$ almost coincides with $\accentset{*}{H}_\mathrm{U}^{(t)}$
and $\widehat{\mathbf{x}}_3^{(4)}$ is quite far away.
The situation with the rotation angle $\Theta_\text{U}$ is similar.
The only exclusion is the observation for the curb's depth $E_\text{U}$ (Fig~\ref{fig:demo-temporal}).
$\widehat{\mathbf{x}}_3^{(4)}$ is evidently closet to the prediction $\accentset{*}{E}_\mathrm{U}^{(t)}$
than $\widehat{\mathbf{x}}_3^{(3)}$ is, but the system cannot make strong inference,
because both candidates are deviated less than $\pm 3 \sigma_{E_\text{U}}$ from the prediction line.
We can conclude that the only inlier is $\widehat{\mathbf{x}}_3^{(3)}$ at time $t$.
Then, $\mathcal{P}_{t}$ is created by updating $\mathcal{P}_{t-1}$ though appending
$\widehat{\mathbf{x}}_3^{(3)}$ and removing the ``oldest'' sample in order to preserve its fixed length.

Demonstration videos of our system can be found in \cite{demoPlaylist}.

\subsection{Detection rate evaluation}

The initial operation on every newly captured camera frame is \textit{curb candidates detection}.
In this section we evaluate the capabilities of our system to correctly detect curbs in the image, before estimating its parameters.
The classification accuracy ($\mathrm{ACC}$) and $F_{1}$ score are measured
by incorporating \textit{Leave-one-video-out} cross validation technique for each individual video sequence in our dataset.
The results are summarized in Table~\ref{tab:classificationRate}.

\begin{table}[t]
	\centering
	\caption{\label{tab:classificationRate}Curb Detection Leave-One-Video-Out Classification Rate.}
	\begin{tabular}{|c|c|c|}
		\hline
		\textbf{Video sequence \#} & \textbf{Accuracy} & \textbf{$\boldsymbol{F}_\mathbf{1}$ score} \\
		\hline
		\hline
		1 & 99.7\% & 0.997 \\
		\hline
		2 & 99.1\% & 0.989 \\
		\hline
		3 & 93.6\% & 0.926 \\
		\hline
		4 & 97.4\% & 0.986 \\
		\hline
		5 & 97.4\% & 0.819 \\
		\hline
		6 & 83.9\% & 0.901 \\	
		\hline
		7 & 96.2\% & 0.974 \\
		\hline
		8 & 81.7\% & 0.871 \\
		\hline
		9 & 90.5\% & 0.940 \\
		\hline
		10 & 91.6\% & 0.956 \\
		\hline
		11 & 81.7\% & 0.798 \\
		\hline
		\hline
		\textbf{Average:} & \textbf{91.4\%} & \textbf{0.923} \\
		\hline
	\end{tabular}
\end{table}

\begin{figure*}[t!]
	\centering
	\def\svgwidth{\textwidth}
	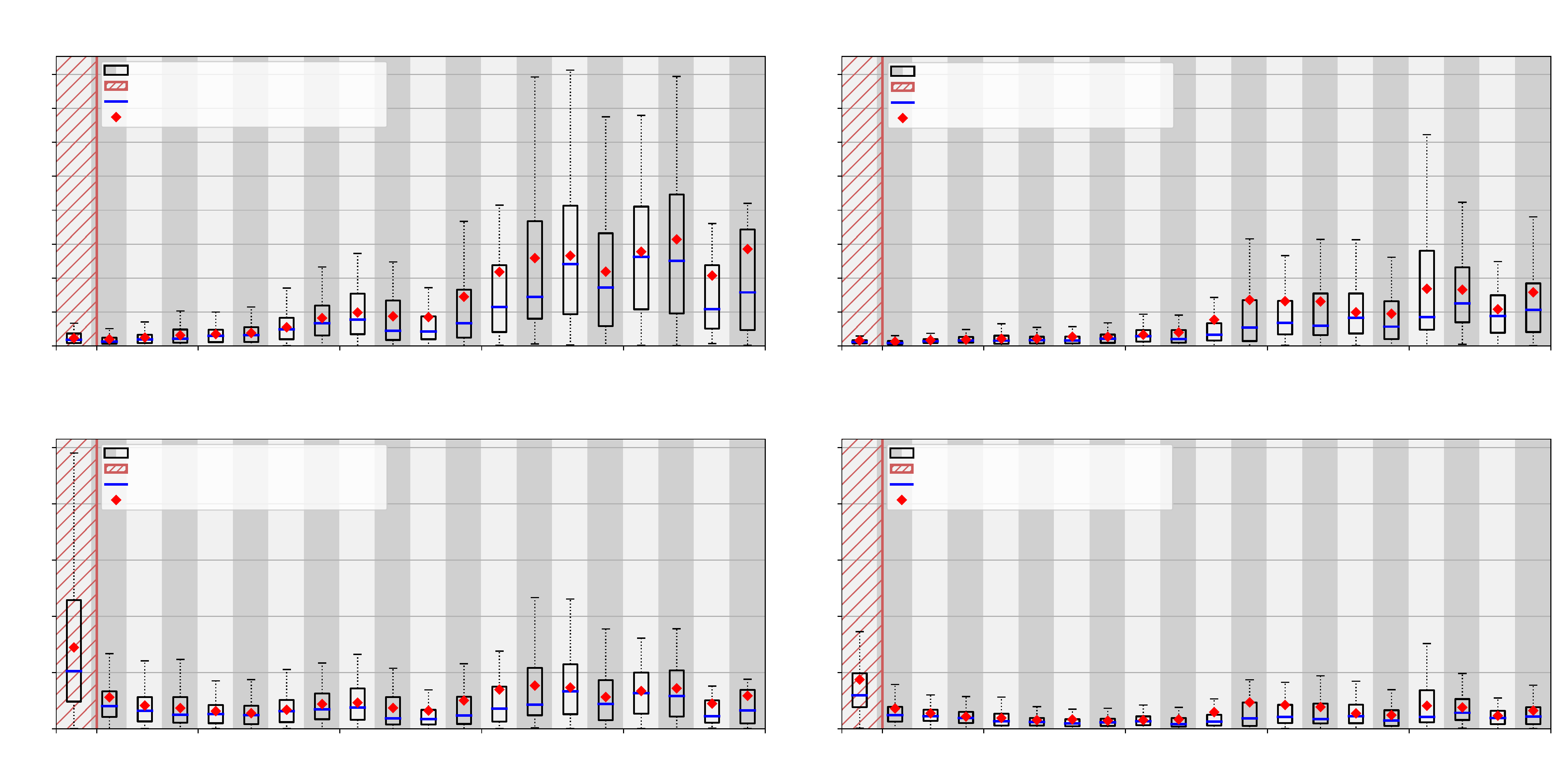
	\caption{
		Absolute (\textit{top}) and relative (\textit{bottom}) curb distance $D_\text{U}$ measurement errors 
		with respect to LIDAR ground truth (GT) data (\textit{left}) and manual labels GT (\textit{right}).}
	\label{fig:result_1-distance}
\end{figure*}

$\mathrm{ACC}$ is calculated according to the following equation
\begin{equation}
\label{eq:classAcc}
  \mathrm{ACC} = \frac{\mathrm{TP} + \mathrm{TN}}{\mathrm{\mathrm{TP} + \mathrm{TN}} + \mathrm{FP} + \mathrm{FN}},
\end{equation}
where $\mathrm{TP}$, $\mathrm{TN}$, $\mathrm{FP}$ and $\mathrm{FN}$ are true positives, true negatives, false positives and false negatives, respectively.
The $F_{1}$ score is given by
\begin{equation}
 F_{1}=2\cdot {\frac {\mathrm {precision} \cdot \mathrm {recall} }{\mathrm {precision} +\mathrm {recall} }},
\end{equation}
where
\begin{equation}
  \mathrm{precision} = \frac{\mathrm{TP}}{\mathrm{TP}+\mathrm{FP}}, \; \mathrm{recall} = \frac{\mathrm{TP}}{\mathrm{TP}+\mathrm{FN}}.
\end{equation}
As can be observed in Table~\ref{tab:classificationRate}, the majority of the videos have accuracy greater than 90\% 
and the average is about 91.40\%. Also $F_{1}$ scores tend to maintain very high values as well.

\subsection{Curb parameters estimation accuracy}

In this section we evaluate system's capabilities as a curb parameters measuring device.
We use two metrics: an absolute~-- \textit{Mean Absolute Error} (MAE) and
a relative~-- \textit{Mean Absolute Percentage Error} (MAPE).
We show how these errors change with respect to the GT distance $D_\text{U}$.
Therefore, we divide the length of CDD in averaging intervals of 25 cm each.
The graphs shown here summarize the results of processing all video sequences from the dataset.

Fig.~\ref{fig:result_1-distance} illustrates box plots of the absolute error of $\widehat{D}_\text{U}$ estimates for our dataset.
$D_\text{U}$ is the only parameter, which has two sources of GT data -- a point-wise LIDAR and manual labels.
Thus, the figure has information for both of them.
The LIDAR labels should be considered as the more accurate baseline measurements,
because they are realized through an independent device, not by the camera.
Hence, the error with respect to them is greater than the one with respect to the manual labels.
Expectedly, the MAE errors and variances are proportional to the distance between the vehicle and the curb.
On the other hand, MAPE errors tend to maintain relatively uniform character through
the entire length of CDD -- below 8-9\%.
The only exception is the closest averaging interval~-- $D_\text{U} \in [0,D_\text{min}]$,
where the curb is partially presented in the camera frame and the system approximates its parameters
using a reduced set of edges (one or two). This can be considered as completely anticipated result.

\begin{figure}
	\centering
	\def\svgwidth{\columnwidth}
	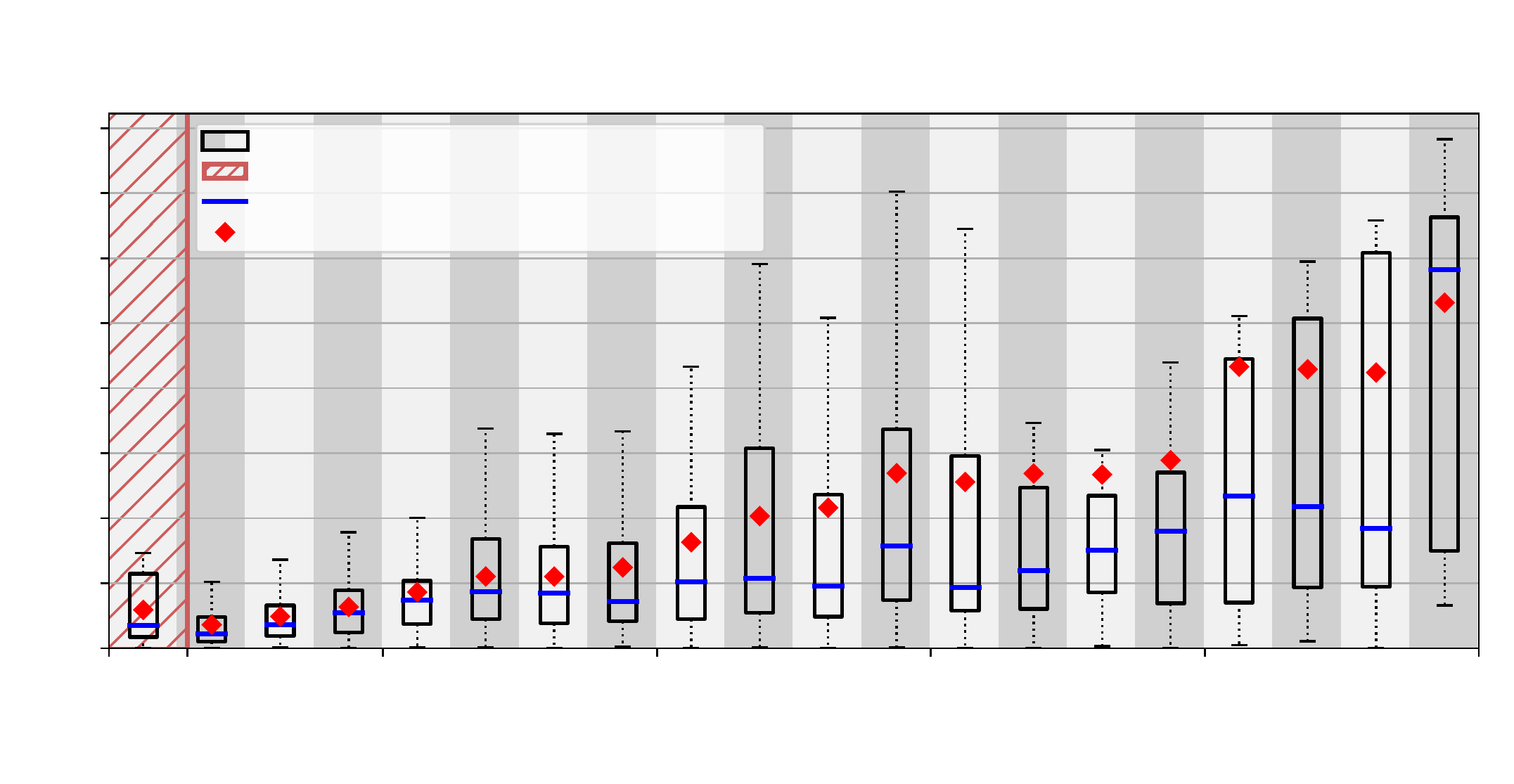
	\caption{Absolute curb rotation angle $\Theta_\text{U}$ measurement errors with respect to the manual labels GT.}
	\label{fig:result_2-rotation}
\end{figure}

Fig.~\ref{fig:result_2-rotation} illustrates the absolute errors of rotation angle measurements with respect to the
manually labeled curb parameters. Similar ot the previous parameter's measurement statistics, the MAE and error variance of
$\Theta_\text{U}$ are proportional to $D_\text{U}$.
Relatively high errors (especially at longer distances) can be explained by the nature of perspective
projection and relative camera and curb positions and orientations. As the camera is relatively close to the road plane,
its optical axis is parallel to it, its field of view is very large (fisheye), large changes in curb rotation are going
to have minimal effect in its appearance in the image.

\begin{figure}
	\centering
	\def\svgwidth{\columnwidth}
	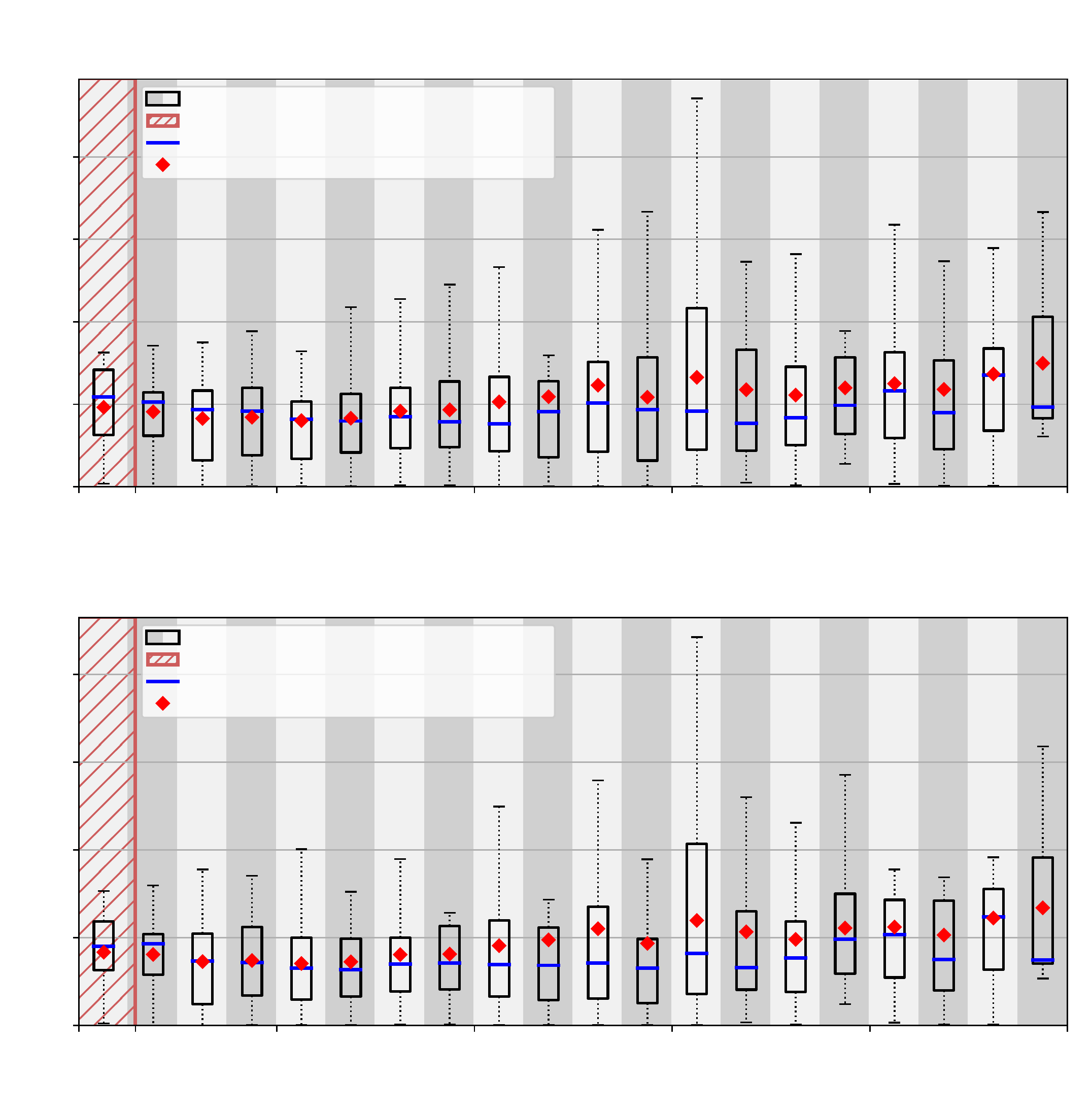
	\caption{Absolute (\textit{top}) and relative (\textit{bottom}) curb height $H_\text{U}$ measurement errors with respect to manual labels GT.}
	\label{fig:result_3-height}
\end{figure}

\begin{figure}
	\centering
	\def\svgwidth{\columnwidth}
	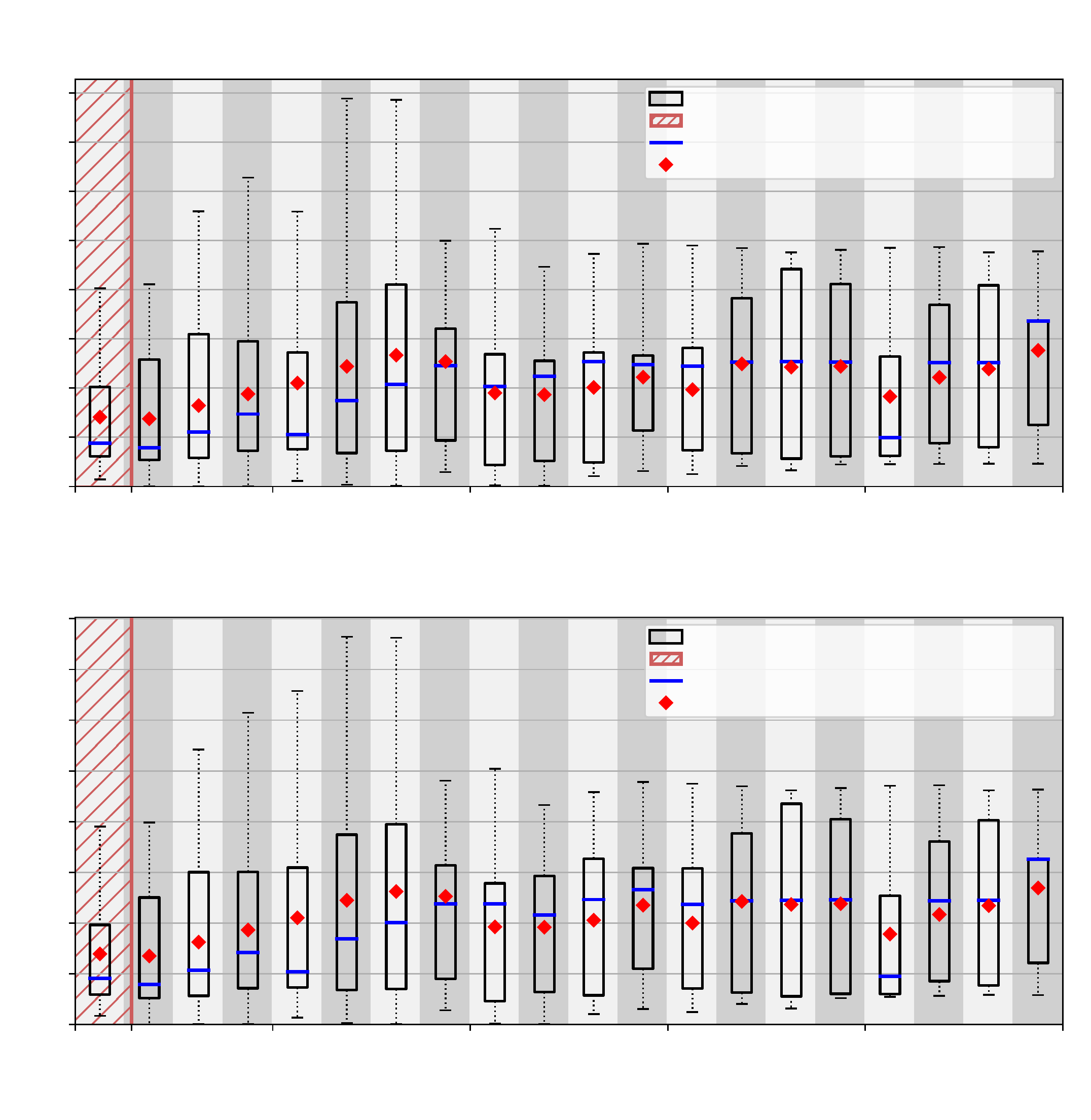
	\caption{Absolute (\textit{top}) and relative (\textit{bottom}) curb depth $E_\text{U}$ measurement errors with respect to manual labels GT.}
	\label{fig:result_4-depth}
\end{figure}

In Fig.~\ref{fig:result_3-height} and Fig.~\ref{fig:result_4-depth} the absolute measurement errors 
of $H_\text{U}$ and $E_\text{U}$ can be seen. The characters of both MAE errors seem to be
relatively uniform, unlike the one of $D_\text{U}$ and $\Theta_\text{U}$. This demonstrates that
these two parameters are less (or not) dependent on the measurement distance $D_\text{U}$.

\section{Conclusion}
\label{sec:conclusions}

In this paper we presented a system for vision-based curb detection and its parameters estimation.
We managed to achieve high detection and curb parameters measurement accuracies just by using a single fisheye camera and CPU computations.
Our algorithm is capable of successfully detecting and tracking curbs at a distance of more than 4 m
with mean absolute error less than $9\%$ for the curb distance estimates in the \textit{Curb Detection Domain}
and not more than $1.5$ cm mean deviation for the curb height estimates.
Robust tracking and high detection rate are accomplished by implementing online temporal analysis.

Here we would like also to discuss the aspects of our future activities, which should be considered as a part of the efforts to improve our system capabilities. One of the main shortcomings is the amount of diversity in the dataset currently used. 
Along with the already available sunny and cloudy daylight conditions, there should be considered also the situations of rain, snow, twilight, night, etc. Another parameter, which should be improved, is the variability of curb types presented in the data. For example, parking curbs or curbs, which consist of rounded edges, may cause problems due to their specific geometry and it is important to test the system performance against it.
As our curb detection system relies on detecting edges in the image, it is also relevant testing the system on various types of the road and sidewalk. For instance, introducing types, which include tiles or pavements.
That will introduce edges, which may confuse the detector.
All these consideration are going to be taken into account when collecting the next datasets.

The current system uses a single camera as an input device and we have proved that it can perform considerably well. 
Other types of sensors are often used as well.
A common practice is the employment of stereo-pairs, which provide depth information as an additional cue.
Another source of 3D data is the LIDARs. Their main drawback is the cost, as it is significantly higher than the one of the cameras. On the other hand, they are active sensors, thus they do not rely on external light sources, such as the sun, street lamps or head lights. Having multiple sources of information requires an appropriate methods to deal with them.
A common approach towards making decision in that cases is by fusing those streams of information. In \cite{7917316} is presented a technique for multi-modal data fusion, which can serve as a method to combine the visual and 3D data for more reliable curb detection.

\appendix[Inverse Perspective-compressing Mapping for~Curb's Scale-invariant Edge Detection]

%

\label{sec:curbScaleInvariantRemapping}

In this section we are presenting the derivation of our method for \textit{Inverse Perspective-compressing Mapping} (IPcM),
whose purpose and application are described in Section~\ref{sec:curbEdgeDetection} and illustrated in Fig.~\ref{fig:curbSizeVsDist}b.
Its objective is minimizing (or even completely eliminating) the dependence of curb's spatial sampling rate $R_\mathrm{U}$
from the distance $D_\mathrm{U}$, 
through its entire definition interval, i.e.\ $R_\mathrm{U}(D_\mathrm{U}) \xrightarrow{\text{remapping}} \widehat{R}_\mathrm{U}(D_\mathrm{U}) = \mathrm{const}: D_\mathrm{U} \in \left [ D_\text{min}, D_\text{max} \right ]$.
The condition to avoid the interpolation is
\begin{equation}
  \widehat{R}_\mathrm{U} \leq \min \left [ R_\mathrm{U}(D_\mathrm{U}) \right ] : {D_\mathrm{U} \in [D_\text{min}, D_\text{max}]}.
\end{equation}
In order to preserve the largest amount of details in the image we take into account the equation (\ref{eq:R_U}) and choose the highest possible value $\widehat{R}_\mathrm{U} = R_\mathrm{U}(D_\text{max})$.

The derivation of the remapping functions presented below consists of two logical parts -- forward and backward (inverse) transforms
\begin{equation}
  \underbrace{(u,v)}_{\substack{\text{original}\\ \text{image space}}} \xrightarrow[\text{transform}]{f \text{ -- forward}} \underbrace{(\widetilde{u}, \widetilde{v})}_{\substack{\text{remapped}\\ \text{image space}}} \xrightarrow[\text{transform}]{\widehat{f} \text{ -- inverse}} \underbrace{(\widehat{u}, \widehat{v})}_{\substack{\text{(pseudo) original}\\ \text{image space}}}
\end{equation}

\subsection{Forward transform (remapping)}

\begin{figure}[t]
\centering
\def\svgwidth{\columnwidth}
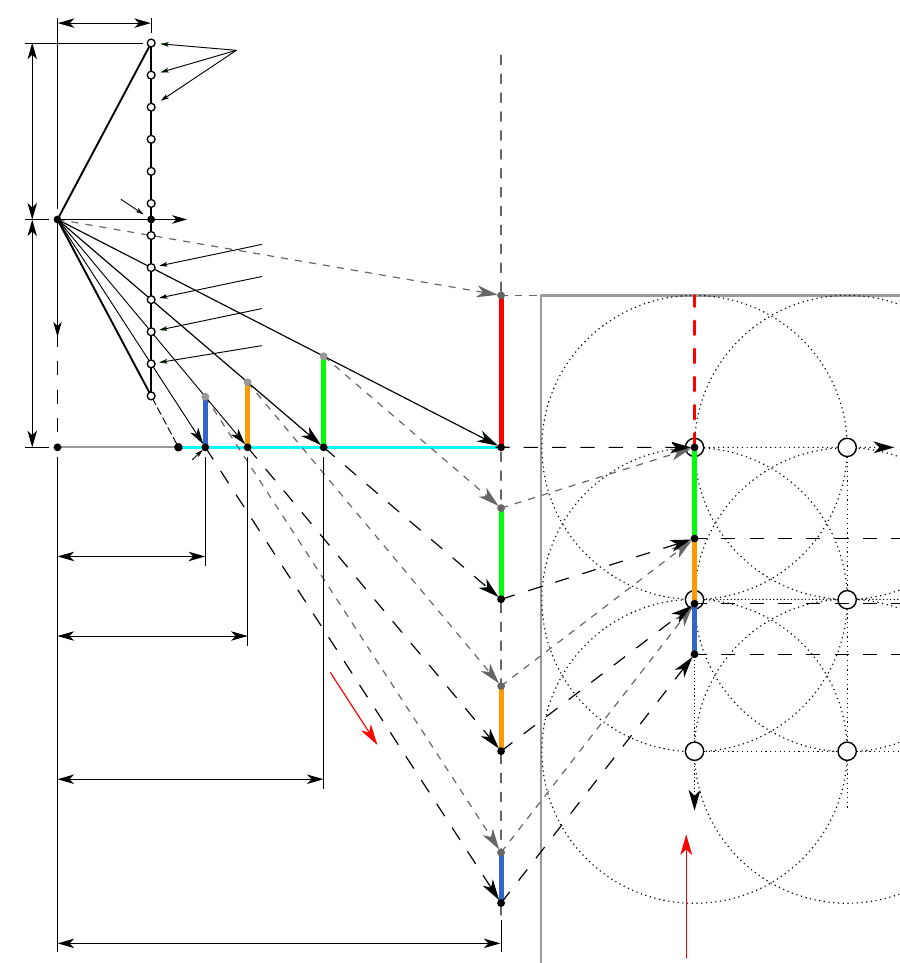
\caption{Curb scale-invariant image remapping: Forward transform -- (vertical) $y$-interpixel distances scaling.}
\label{fig:scaleInvariant1}
\end{figure}

This transformation defines the mapping of the image point (pixel) coordinates from the original image to the new one.
It is important to note that we consider pixels as sampling locations. i.e.\ points from the camera image plane with no designated areas, contrary to the common understanding that they are small regions.
Thus, the remapping procedure in essence is scaling of the distances between the pixels in an appropriate way. Moreover, the camera's pixels grid is regular and square, i.e.\ all the inter-pixel distances are equal.

Fig.~\ref{fig:scaleInvariant1} depicts the theoretical foundations of our approach.
Every pixel from image plane defines a ray which originates in $O_\mathrm{C}$ -- camera's coordinate frame origin, and passes through its location on the image plane.
We will refer to these rays as \textit{sampling rays} and their intersection points with the road plane -- \textit{sampling points}.
Furthermore, due to our system's geometrical setup, all sampling points corresponding to the pixels sharing the same row from the image will have the identical $z_\mathrm{C}$ coordinates, i.e.\ they will be equidistant w.r.t. the camera's coordinate plane $x_\mathrm{C} y_\mathrm{C}$. Thus, by ``a sampling ray" or ``a sampling point" we will refer to all the rays and points corresponding to the pixels from the same image row.
 
Let $y_i$ be a positive vertical coordinate of the pixels located on the $v_i$-th image row, i.e.\ $v_i > c_y$, where $v_i = y_i + c_y$ and $i$ is row index.
Hence, $y_i$ will correspond to a sampling ray which intersects the road plane in the sampling point $\mathbf{P}_i \in \mathbb{R}^3$ at a distance $D_i$ that can be calculated by the equation (\ref{eq:measuringDistances}).
Let $D = \{D_i\}$ be a set, whose members are all the possible distances $D_i$ defined by the camera.
Also we define the longest sampling distance $\widehat{D}_\text{max}$, where $\widehat{D}_\text{max} = \sup \left \{ D_j \in D : D_j \leq D_\text{max} \right \}$.
For the sake of simplicity we set the index $i = 0$ for the image pixel row which corresponds to the distance $\widehat{D}_\text{max}$, i.e.\ $y_0 \leftrightarrow \mathbf{P}_0 \leftrightarrow D_0 = \widehat{D}_\text{max}$.

Let $s_i$ be the distance between the two sampling rays corresponding to the pixel rows $y_i$ and  $y_{i-1}$ along vertical direction, measured at the sampling point $\mathbf{P}_i$ (Fig.~\ref{fig:scaleInvariant1}). Then we get
\begin{equation}
  s_i = \frac{D_i}{f_y} = \frac{1}{R_\mathrm{U}(D_i)},
\end{equation}
which shows that $s_i \propto D_i$. Hence, normalizing $R_\mathrm{U}$ for all sampling positions $D_i$, which is our goal, is equivalent to normalizing $s_i$ and we do it by using the distance $s_0$ at $D_0$ as a reference and define the scaling factors $S_i$, such that
\begin{equation}
  S_{i} = \frac{s_{i}}{s_{0}} = \frac{D_i}{D_0} = \frac{y_0}{y_i}.
\end{equation}

On Fig.~\ref{fig:scaleInvariant1} we can see the geometrical interpretation of the scaling process.
Let's imagine that a curb with height $H_\mathrm{U}$ slides along the axis $z_\mathrm{C}$ on the ground. Hence, the height of its projection in the image is going to change inversely proportional to the distance $D_\mathrm{U}$ (as we have already seen on Fig.~\ref{fig:curbSizeVsDist}a), i.e.\ the farther the curb is, the smaller its projection in the camera will be. We want to resolve that problem by making curbs size in the image fixed for all the distances $D_\mathrm{U} \in [D_\text{min}, D_\text{max}]$.
Our approach to achieve it is virtually "dragging" the curb along the sampling ray corresponding to the point $\mathbf{P}_i$ from its location at the distance $D_\mathrm{U} = D_i$ to a plane which is parallel to $x_\mathrm{C}y_\mathrm{C}$ and passes trough the point $\mathbf{P}_0$.
Thus, the curb basically will always be at the fixed distance $D_\mathrm{U} = D_0$ w.r.t. the camera, thereby its size in the image will be constant, and at the same time it will still be projected in the image at the vertical position $y_i$ which corresponds to the point of its real location $D_i$. 

Since we do not use any of the image information outside it's RoI, there is no need to spend processing time on remapping it. 
That's why, we choose that the $y_0$ is mapped to the first raw of the resulting image, i.e.\ $\widetilde{v}_0 = f_v(y_0) = 0 \; \text{px}$, where $f_v$ is the remapping function if the vertical point coordinates only. Based on Fig.~\ref{fig:scaleInvariant1} for $f_v$ we can write
\begin{equation}
    \widetilde{v}_{n} = f_v(y_n) = \sum_{i = 1}^{n} S_{i} = \sum_{i = 1}^{n} \frac{y_{0}}{y_{i}} 
                   = y_{0} \sum_{i = 1}^{n} \frac{1}{y_{0} + i}.
\end{equation}
Obviously, $f_v$ has an opened form, which is inconvenient to estimate its inverse. Fortunately, the sum term has the pattern of harmonic series, with the difference that the sum is not infinite. However, it can be presented by a difference of two harmonic series partial sums
\begin{equation}
  \label{eq:f_v_approx}
  f_v(y_n) \approx y_{0} \left ( \sum_{k=1}^{\widetilde{y}_n} \frac{1}{k} - \sum_{k=1}^{\widetilde{y}_0 } \frac{1}{k} \right ) \approx y_{0} \left ( H_{{y}_n} - H_{{y}_0} \right ),
\end{equation}
where
\begin{equation}
  \label{eq:harmonicNumber}
  H_q = \ln(q) + \gamma + \frac{1}{2q} - \frac{1}{12q^2} + \frac{1}{120 q^4} - \varepsilon_q
\end{equation}
is the $q$-th harmonic number determined by expanding some of the terms in the \textit{Hurwitz zeta function}, $\gamma$ is the \textit{Euler-Mascheroni constant} and $\varepsilon_q \in \left ( 0, \frac{1}{252q^6} \right )$.
$\widetilde{y}_n$ and $\widetilde{y}_0$ are the floored values of ${y}_n$ and ${y}_0$, respectively.
In the general case, $c_x$ and $c_y$ -- the coordinates of the camera's principal point $\mathbf{c}$, are real numbers.
Hence, $y_n$ and $y_0$ will also be real which cannot be used as a limit of the sum operators in (\ref{eq:f_v_approx}).
Therefore, we floor them by taking into account that the produced error will be negligible.
After the subtraction of $H_{{y}_n}$ and $H_{{y}_0}$, $\gamma$ is eliminated
and the influence of the last 3 terms in (\ref{eq:harmonicNumber}) from practical point of view is imperceptible.
The term $\frac{1}{2q}$ is very well approximated by $\ln{ \left ( \frac{q}{q-\frac{1}{2}} \right )}$, estimated by the \textit{Laurent series} expansion.
The estimated total approximation error is less than $2\cdot10^{-3}$ pixels and we get the final simplified closed form expression for $f_v$
\begin{equation}
  \label{eq:f_v}
  \widetilde{v} = f_v(v) = y_0 \ln \left [ \frac{(v - c_y)^2}{y_0(v - c_y-\frac{1}{2})} \right ] - \frac{1}{2}
\end{equation}
Every line of the image is scaled symmetrically with the same scale factor $S_i$, i.e.\ the mapping function for the horizontal direction has the following form
\begin{equation}
  \label{eq:f_u}
  \widetilde{u} = f_u(u, v) = y_0 \frac{u - c_x}{v-c_y} + c_x
\end{equation}

\subsection{Inverse transform}
To derive the inverse transform we just inverse equations (\ref{eq:f_v}) and (\ref{eq:f_u}) and we get
\begin{equation}
  \begin{split}
    \widehat{v} &= g_v(\widetilde{v}) = \frac{m + \sqrt{m(m -2)}}{2} + c_y \\  
    \widehat{u} &= g_u(\widetilde{u}, \widehat{v}) = \frac{(\widetilde{u} - c_x)(\widehat{v}-c_y)}{y_0} + c_x
  \end{split},
\end{equation}
where
\begin{equation}
  m = y_0 \exp \left ( \frac{\widetilde{v} + \frac{1}{2}}{y_0} \right ).
\end{equation}


\section*{Acknowledgment}

We would like to thank the anonymous reviewers for their valuable comments,
which allowed us to improve the presentation of our paper and resolve various imperfections.
This work was partially supported by General Motors Research.
Stanislav Panev's scholarship in CMU was funded by the Bulgarian-American Commission of Education Exchange ``Fulbright'' with a grant number 15-21-04 from June 3, 2015.

\ifCLASSOPTIONcaptionsoff
  \newpage
\fi



\bibliographystyle{IEEEtran}
\bibliography{IEEEabrv,bibliography,bibliography-RelatedWork}

\begin{thebibliography}{10}
\providecommand{\url}[1]{#1}
\csname url@samestyle\endcsname
\providecommand{\newblock}{\relax}
\providecommand{\bibinfo}[2]{#2}
\providecommand{\BIBentrySTDinterwordspacing}{\spaceskip=0pt\relax}
\providecommand{\BIBentryALTinterwordstretchfactor}{4}
\providecommand{\BIBentryALTinterwordspacing}{\spaceskip=\fontdimen2\font plus
\BIBentryALTinterwordstretchfactor\fontdimen3\font minus
  \fontdimen4\font\relax}
\providecommand{\BIBforeignlanguage}[2]{{%
\expandafter\ifx\csname l@#1\endcsname\relax
\typeout{** WARNING: IEEEtran.bst: No hyphenation pattern has been}%
\typeout{** loaded for the language `#1'. Using the pattern for}%
\typeout{** the default language instead.}%
\else
\language=\csname l@#1\endcsname
\fi
#2}}
\providecommand{\BIBdecl}{\relax}
\BIBdecl

\bibitem{zhao_curb_2012}
G.~Zhao and J.~Yuan, ``Curb detection and tracking using 3d-{LIDAR} scanner,''
  in \emph{2012 19th {IEEE} {International} {Conference} on {Image}
  {Processing}}, Sep. 2012, pp. 437--440.

\bibitem{yao_road_2012}
W.~Yao, Z.~Deng, and L.~Zhou, ``Road curb detection using 3d {LIDAR} and
  integral laser points for intelligent vehicles,'' in \emph{The 6th
  {International} {Conference} on {Soft} {Computing} and {Intelligent}
  {Systems}, and {The} 13th {International} {Symposium} on {Advanced}
  {Intelligence} {Systems}}, Nov. 2012, pp. 100--105.

\bibitem{hata_robust_2014}
A.~Y. Hata, F.~S. Osorio, and D.~F. Wolf, ``Robust curb detection and vehicle
  localization in urban environments,'' in \emph{2014 {IEEE} {Intelligent}
  {Vehicles} {Symposium} {Proceedings}}, Jun. 2014, pp. 1257--1262.

\bibitem{chen_velodyne-based_2015}
T.~Chen, B.~Dai, D.~Liu, J.~Song, and Z.~Liu, ``Velodyne-based curb detection
  up to 50 meters away,'' in \emph{2015 {IEEE} {Intelligent} {Vehicles}
  {Symposium} ({IV})}, Jun. 2015, pp. 241--248.

\bibitem{zhang_real-time_2015}
Y.~Zhang, J.~Wang, X.~Wang, C.~Li, and L.~Wang, ``A real-time curb detection
  and tracking method for {UGVs} by using a 3d-{LIDAR} sensor,'' in \emph{2015
  {IEEE} {Conference} on {Control} {Applications} ({CCA})}, Sep. 2015, pp.
  1020--1025.

\bibitem{scheunert_free_2007}
U.~Scheunert, B.~Fardi, N.~Mattern, G.~Wanielik, and N.~Keppeler, ``Free space
  determination for parking slots using a 3d {PMD} sensor,'' in \emph{2007
  {IEEE} {Intelligent} {Vehicles} {Symposium}}, Jun. 2007, pp. 154--159.

\bibitem{gallo_robust_2008}
O.~Gallo, R.~Manduchi, and A.~Rafii, ``Robust curb and ramp detection for safe
  parking using the {Canesta} {TOF} camera,'' in \emph{2008 {IEEE} {Computer}
  {Society} {Conference} on {Computer} {Vision} and {Pattern} {Recognition}
  {Workshops}}, Jun. 2008, pp. 1--8.

\bibitem{byun_autonomous_2011}
J.~Byun, J.~Sung, M.~C. Roh, and S.~H. Kim, ``Autonomous driving through {Curb}
  detection and tracking,'' in \emph{2011 8th {International} {Conference} on
  {Ubiquitous} {Robots} and {Ambient} {Intelligence} ({URAI})}, Nov. 2011, pp.
  273--277.

\bibitem{liu_curb_2013}
Z.~Liu, D.~Liu, T.~Chen, and C.~Wei, ``Curb {Detection} {Using} 2d {Range}
  {Data} in a {Campus} {Environment},'' in \emph{2013 {Seventh} {International}
  {Conference} on {Image} and {Graphics}}, Jul. 2013, pp. 291--296.

\bibitem{pollard_step_2013}
E.~Pollard, J.~Perez, and F.~Nashashibi, ``Step and curb detection for
  autonomous vehicles with an algebraic derivative-based approach applied on
  laser rangefinder data,'' in \emph{2013 {IEEE} {Intelligent} {Vehicles}
  {Symposium} ({IV})}, Jun. 2013, pp. 684--689.

\bibitem{oniga_curb_2008}
F.~Oniga, S.~Nedevschi, and M.~M. Meinecke, ``Curb {Detection} {Based} on a
  {Multi}-{Frame} {Persistence} {Map} for {Urban} {Driving} {Scenarios},'' in
  \emph{2008 11th {International} {IEEE} {Conference} on {Intelligent}
  {Transportation} {Systems}}, Oct. 2008, pp. 67--72.

\bibitem{siegemund_temporal_2011}
J.~Siegemund, U.~Franke, and W.~Förstner, ``A temporal filter approach for
  detection and reconstruction of curbs and road surfaces based on
  {Conditional} {Random} {Fields},'' in \emph{2011 {IEEE} {Intelligent}
  {Vehicles} {Symposium} ({IV})}, Jun. 2011, pp. 637--642.

\bibitem{oniga_curb_2011}
F.~Oniga and S.~Nedevschi, ``Curb detection for driving assistance systems: {A}
  cubic spline-based approach,'' in \emph{2011 {IEEE} {Intelligent} {Vehicles}
  {Symposium} ({IV})}, Jun. 2011, pp. 945--950.

\bibitem{kellner_multi-cue_2015}
M.~Kellner, U.~Hofmann, M.~E. Bouzouraa, and N.~Stephan, ``Multi-cue,
  {Model}-{Based} {Detection} and {Mapping} of {Road} {Curb} {Features} {Using}
  {Stereo} {Vision},'' in \emph{2015 {IEEE} 18th {International} {Conference}
  on {Intelligent} {Transportation} {Systems}}, Sep. 2015, pp. 1221--1228.

\bibitem{fernandez_curvature-based_2015}
C.~Fernández, D.~F. Llorca, C.~Stiller, and M.~A. Sotelo, ``Curvature-based
  curb detection method in urban environments using stereo and laser,'' in
  \emph{2015 {IEEE} {Intelligent} {Vehicles} {Symposium} ({IV})}, Jun. 2015,
  pp. 579--584.

\bibitem{haltakov_scene_2012}
V.~Haltakov, H.~Belzner, and S.~Ilic, ``Scene understanding from a moving
  camera for object detection and free space estimation,'' in \emph{2012 {IEEE}
  {Intelligent} {Vehicles} {Symposium}}, Jun. 2012, pp. 105--110.

\bibitem{seibert_camera_2013}
A.~Seibert, M.~Hähnel, A.~Tewes, and R.~Rojas, ``Camera based detection and
  classification of soft shoulders, curbs and guardrails,'' in \emph{2013
  {IEEE} {Intelligent} {Vehicles} {Symposium} ({IV})}, Jun. 2013, pp. 853--858.

\bibitem{prinet_3d_2016}
V.~Prinet, J.~Wang, J.~Lee, and D.~Wettergreen, ``3d road curb extraction from
  image sequence for automobile parking assist system,'' in \emph{2016 {IEEE}
  {International} {Conference} on {Image} {Processing} ({ICIP})}, Sep. 2016,
  pp. 3847--3851.

\bibitem{ViolaJones}
P.~Viola and M.~Jones, ``Rapid object detection using a boosted cascade of
  simple features,'' in \emph{Proceedings of the 2001 IEEE Computer Society
  Conference on Computer Vision and Pattern Recognition. CVPR 2001}, vol.~1,
  2001, pp. I--511--I--518 vol.1.

\bibitem{fisheyeCalibration}
J.~Kannala and S.~S. Brandt, ``A generic camera model and calibration method
  for conventional, wide-angle, and fish-eye lenses,'' \emph{IEEE Transactions
  on Pattern Analysis and Machine Intelligence (PAMI)}, vol.~28, no.~8, pp.
  1335--1340, Aug 2006.

\bibitem{HoughTransform}
R.~O. Duda and P.~E. Hart, ``Use of the hough transformation to detect lines
  and curves in pictures,'' \emph{Commun. ACM}, vol.~15, no.~1, pp. 11--15,
  Jan. 1972.

\bibitem{Canny}
J.~Canny, ``A computational approach to edge detection,'' \emph{IEEE Trans.
  Pattern Anal. Mach. Intell.}, vol.~8, no.~6, pp. 679--698, Jun. 1986.

\bibitem{Sobel1968}
I.~Sobel and G.~Feldman, ``{A 3x3 Isotropic Gradient Operator for Image
  Processing},'' 1968, never published but presented at a talk at the Stanford
  Artificial Project.

\bibitem{duda1973pattern}
R.~Duda and P.~Hart, \emph{Pattern Classification and Scene Analysis}.\hskip
  1em plus 0.5em minus 0.4em\relax Wiley, 1973, pp. 271--272.

\bibitem{jähne1999handbook}
B.~J{\"a}hne, H.~Scharr, and S.~K{\"o}rkel, ``Principles of filter design,'' in
  \emph{B. J{\"a}hne, H. Haussecker and P. Geissler, Handbook of Computer
  Vision and Applications}.\hskip 1em plus 0.5em minus 0.4em\relax Academic
  Press, 1999, no. v. 2, pp. 125--152.

\bibitem{lipkin1970picture}
J.~Prewitt, ``Object enhancement and extraction,'' in \emph{B.S. Lipkin and A.
  Rosenfeld, editors, Picture Processing and Psychopictorics}.\hskip 1em plus
  0.5em minus 0.4em\relax Academic Press, 1970, ch.~1, pp. 75--150.

\bibitem{Roberts63}
L.~G. Roberts, \emph{Machine Perception of Three-Dimensional Solids}, ser.
  Outstanding Dissertations in the Computer Sciences.\hskip 1em plus 0.5em
  minus 0.4em\relax Garland Publishing, New York, 1963.

\bibitem{LMA1}
K.~Levenberg, ``A method for the solution of certain non-linear problems in
  least squares,'' \emph{Quarterly of Applied Mathematics}, vol.~2, no.~2, pp.
  164--168, 1944.

\bibitem{LMA2}
D.~W. Marquardt, ``An algorithm for least-squares estimation of nonlinear
  parameters,'' \emph{Journal of the Society for Industrial and Applied
  Mathematics}, vol.~11, no.~2, pp. 431--441, 1963.

\bibitem{HOG}
N.~Dalal and B.~Triggs, ``Histograms of oriented gradients for human
  detection,'' in \emph{2005 IEEE Computer Society Conference on Computer
  Vision and Pattern Recognition (CVPR'05)}, vol.~1, June 2005, pp. 886--893
  vol. 1.

\bibitem{liblinear}
\BIBentryALTinterwordspacing
R.-E. Fan, K.-W. Chang, C.-J. Hsieh, X.-R. Wang, and C.-J. Lin, ``{LIBLINEAR}:
  A library for large linear classification,'' \emph{Journal of Machine
  Learning Research}, vol.~9, pp. 1871--1874, 2008. [Online]. Available:
  \url{http://www.csie.ntu.edu.tw/~cjlin/liblinear}
\BIBentrySTDinterwordspacing

\bibitem{opencv_library}
G.~Bradski, ``{The OpenCV Library},'' \emph{Dr. Dobb's Journal of Software
  Tools}, 2000.

\bibitem{demoPlaylist}
``Curb detection system demonstration videos,'' \url{https://goo.gl/NdMJFY},
  accessed: 2017-04-24.

\bibitem{7917316}
H.~Liu, Y.~Wu, F.~Sun, B.~Fang, and D.~Guo, ``Weakly paired multimodal fusion
  for object recognition,'' \emph{IEEE Transactions on Automation Science and
  Engineering}, vol.~15, no.~2, pp. 784--795, April 2018.

\end{thebibliography}

%
%

%


\begin{IEEEbiography}[{\includegraphics[width=1in,height=1.25in,clip,keepaspectratio]{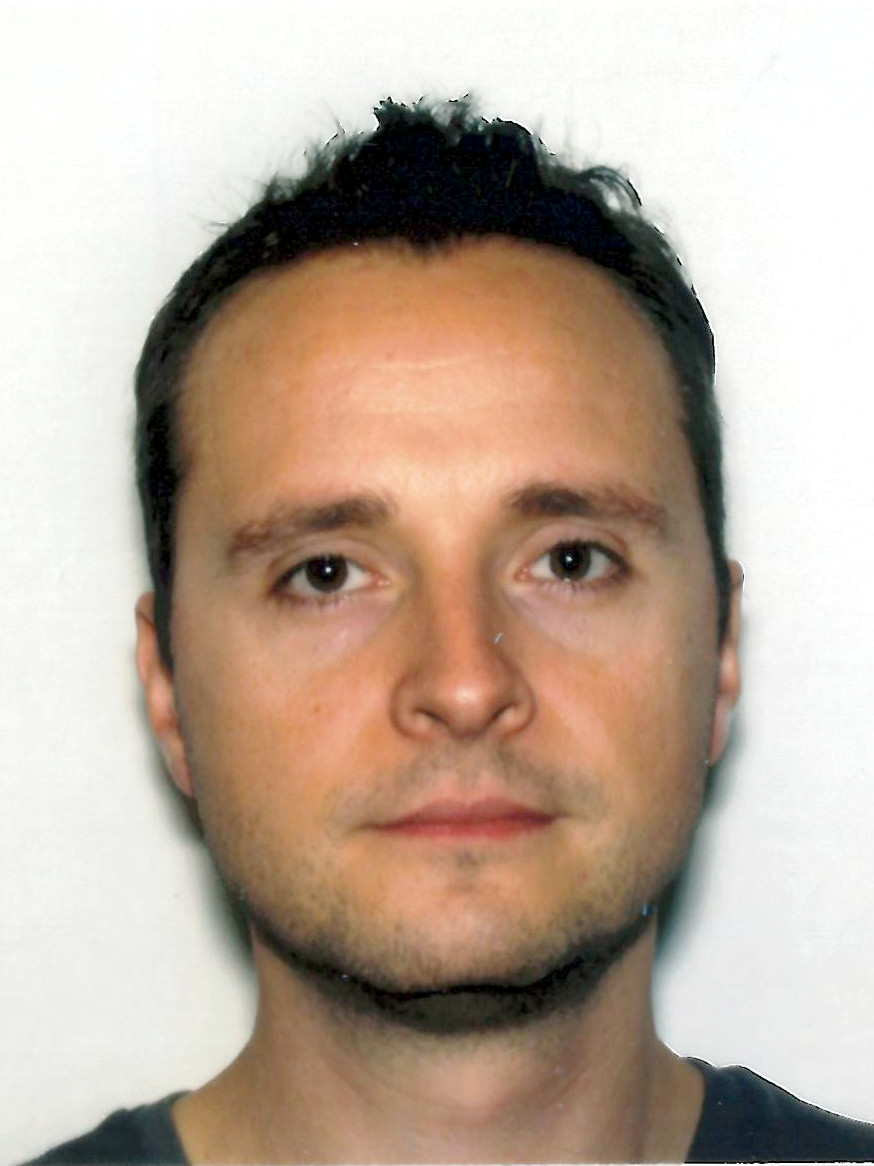}}]{Stanislav Panev}
is currently a postdoctoral fellow at the Robotics Institute of Carnegie Mellon University.
He received BSc and MSc degrees in telecommunications, as well as PhD degrees in computer science from the Technical University of Sofia, Bulgaria in 2005, 2008 and 2014, respectively.
His research topics include, but are not limited to: computer vision, machine learning, computer graphics, artificial intelligence.
\end{IEEEbiography}

\begin{IEEEbiography}[{\includegraphics[width=1in,height=1.25in,clip,keepaspectratio]{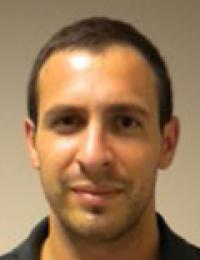}}]{Francisco Vicente}
received a B.Sc. degree in tele communications and a M.Sc. degree in telematics from Universidad Politecnica de Cartagena, Cartagena, Spain, in 2007 and 2008, respectively. In addition, he received a M.S. degree in Robotics at Carnegie Mellon University, Pittsburgh, PA, USA. in 2015.

Since 2015, he has been a Scientific Specialist with the Robotics Institute, Carnegie Mellon University. His research interests are in the fields of computer vision and machine learning.
\end{IEEEbiography}

\vfill

\newpage
 
\begin{IEEEbiography}[{\includegraphics[width=1in,height=1.25in,clip,keepaspectratio]{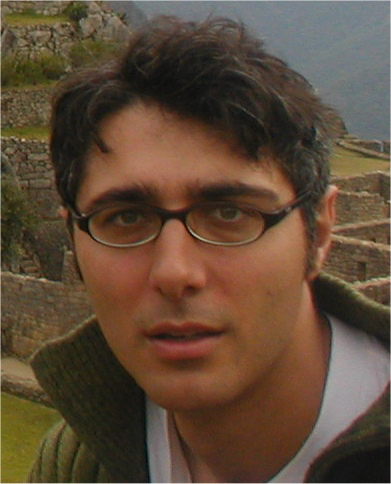}}]{Fernando De la Torre}
received the BSc degree in telecommunications, as well as the MSc and PhD degrees in electronic engineering from the La~Salle School of Engineering at Ramon Llull University, Barcelona, Spain in 1994, 1996, and 2002, respectively. He is an associate research professor in the Robotics Institute at Carnegie Mellon University. His research interests are in the fields of computer vision and machine learning. Currently, he is directing the Component Analysis Laboratory (http://ca.cs.cmu.edu) and the Human Sensing Laboratory (http://humansensing.cs.cmu.edu) at Carnegie Mellon University. He has more than 150 publications in referred journals and conferences and is an associate editor at IEEE TPAMI. He has organized and co-organized several workshops and has given tutorials at international conferences on component analysis.
\end{IEEEbiography}

\begin{IEEEbiography}[{\includegraphics[width=1in,height=1.25in,clip,keepaspectratio]{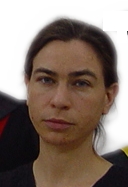}}]{V\'{e}ronique Prinet}
	is currently a Visiting Professor at The Hebrew University of Jerusalem, Israel.
	She received a Ph.D. in Computer Science from INRIA, in collaboration with University Paris-11 Orsay and CNES (French National Center for Spatial Studies).
	In 1999, she was a post-doctoral research fellow at the Institute of Automation, Chinese Academy of Sciences (CASIA), Beijing.
	She subsequently hold an Associate Professor position in the same institution from 2000 to 2012.
	She worked as a senior researcher at General Motors Advanced Technical Center, Israel, from 2014 to 2016.
	Her research interests are in data-driven approaches to image processing and computer vision.
\end{IEEEbiography}


\vfill


\end{document}